%% file: main.tex
\documentclass[10pt,journal,compsoc]{IEEEtran}
\ifCLASSOPTIONcompsoc
  \usepackage[nocompress]{cite}
\else
  \usepackage{cite}
\fi
\ifCLASSINFOpdf
\else
\fi
\usepackage{amsmath}
\usepackage{amssymb}
\usepackage{algorithm}
\usepackage{graphicx}
\usepackage{subfigure}
\usepackage{amsthm}

\newtheorem{assumption}{Assumption}
\newtheorem{definition}{Definition}
\newtheorem{theorem}{Theorem}
\newtheorem{lemma}{Lemma}
\newtheorem{problem}{Problem}
\newtheorem{remark}{Remark}
\usepackage{algorithmic}

\usepackage{array}
\usepackage{url}

\begin{document}
%
\title{Distributionally Robust Learning with Stable Adversarial Training}
%
%
%
%

\author{Jiashuo~Liu, Zheyan~Shen, Peng~Cui\footnote{},\IEEEmembership{Senior Member,~IEEE}, Linjun~Zhou, Kun~Kuang, Bo~Li
\IEEEcompsocitemizethanks{\IEEEcompsocthanksitem J. Liu, Z. Shen, P. Cui and L. Zhou are with the Department
of Computer Science and Technology, Tsinghua University, Beijing, China. B. Li is with the School of Economics and Management, Tsinghua University, Beijing, China. K. Kuang is with the College of Computer Science and Technology, Zhejiang University, Hangzhou, China.\protect\\
E-mail: liujiashuo77@gmail.com, shenzy13@qq.com, cuip@tsinghua.edu.cn, zhoulj16@mails.tsinghua.edu.cn, kunkuang@zju.edu.cn, libo@sem.tsinghua.edu.cn.\protect\\ 
\IEEEcompsocthanksitem Peng Cui and Bo Li are the corresponding authors.
}

}

\IEEEtitleabstractindextext{%
\begin{abstract}
Machine learning algorithms with empirical risk minimization are vulnerable under distributional shifts due to the greedy adoption of all the correlations found in training data. 
There is an emerging literature on tackling this problem by minimizing the worst-case risk over an uncertainty set. 
However, existing methods mostly construct ambiguity sets by treating all variables equally regardless of the stability of their correlations with the target, resulting in the overwhelmingly-large uncertainty set and low confidence of the learner.
In this paper, we propose a novel Stable Adversarial Learning (SAL) algorithm that leverages heterogeneous data sources to construct a more practical uncertainty set and conduct differentiated robustness optimization, where covariates are differentiated according to the stability of their correlations with the target. 
We theoretically show that our method is tractable for stochastic gradient-based optimization and provide the performance guarantees for our method. 
Empirical studies on both simulation and real datasets validate the effectiveness of our method in terms of uniformly good performance across unknown distributional shifts.
\end{abstract}

}

\maketitle

\IEEEdisplaynontitleabstractindextext

%
\IEEEpeerreviewmaketitle

\input{data/1intro}

\input{data/2related}

\input{data/3problem}

\input{data/4method}

\input{data/5theory}

\input{data/6experiments}

\section{Conclusion}
In this paper, we address a practical problem of overwhelmingly-large uncertainty set in robust learning, which often results in unsatisfactory performance under distributional shifts in real situations. 
We propose the Stable Adversarial Learning (SAL) algorithm that anisotropically considers each covariate to achieve more realistic robustness. 
We theoretically show that our method constructs a better uncertainty set and provide the theoretical guarantee for our method.
Empirical studies validate the effectiveness of our methods in terms of uniformly good performance across different distributed data.

\section*{Acknowledgments}
We would like to thank the anonymous reviewers for their constructive suggestions and efforts to improve this paper.
This work was supported in part by National Key R\&D Program of China (No. 2018AAA0102004), National Natural Science Foundation of China (No. U1936219, 62141607), Beijing Academy of Artificial Intelligence (BAAI). 
Kun Kuang's research was supported by National Natural Science Foundation of China (U20A20387, 62006207), National Key Research and Development Program of China (2021YFC3340300), Young Elite Scientists Sponsorship Program by CAST (2021QNRC001)
Bo Li’s research was supported by the National Natural Science Foundation of China (No.72171131); the Tsinghua University Initiative Scientific Research Grant (No. 2019THZWJC11); Technology and Innovation Major Project of the Ministry of Science and Technology of China under Grants 2020AAA0108400 and 2020AAA0108403.


%

%
%
%
%
%

\ifCLASSOPTIONcaptionsoff
  \newpage
\fi





\bibliographystyle{IEEEtran}
\bibliography{IEEEabrv,tkde}
%

%

\begin{IEEEbiography}[{\includegraphics[width=1.0in,height=1.25in,keepaspectratio]{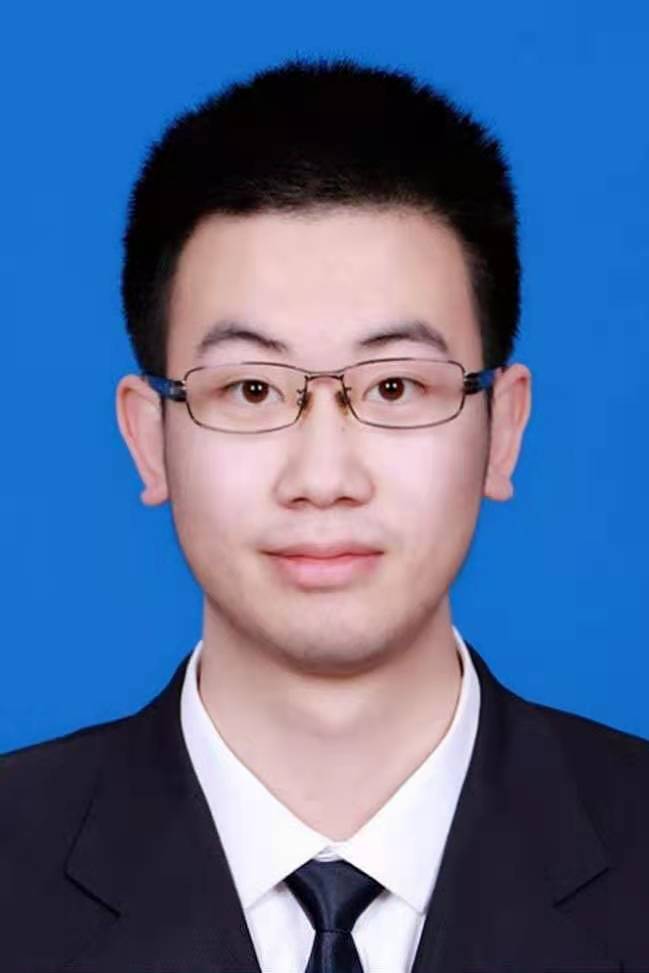}}]{Jiashuo Liu}
received his BE degree from the Department of Computer Science and Technology, Tsinghua University in 2020. He is currently pursuing a Ph.D. Degree in the Department of Computer Science and Technology at Tsinghua University. His research interests focus on causally-regularized machine learning and distributionally robust learning, especially in developing algorithms with stable performance under distributional shifts. 
\end{IEEEbiography}

\begin{IEEEbiography}[{\includegraphics[width=1.0in,height=1.25in,keepaspectratio]{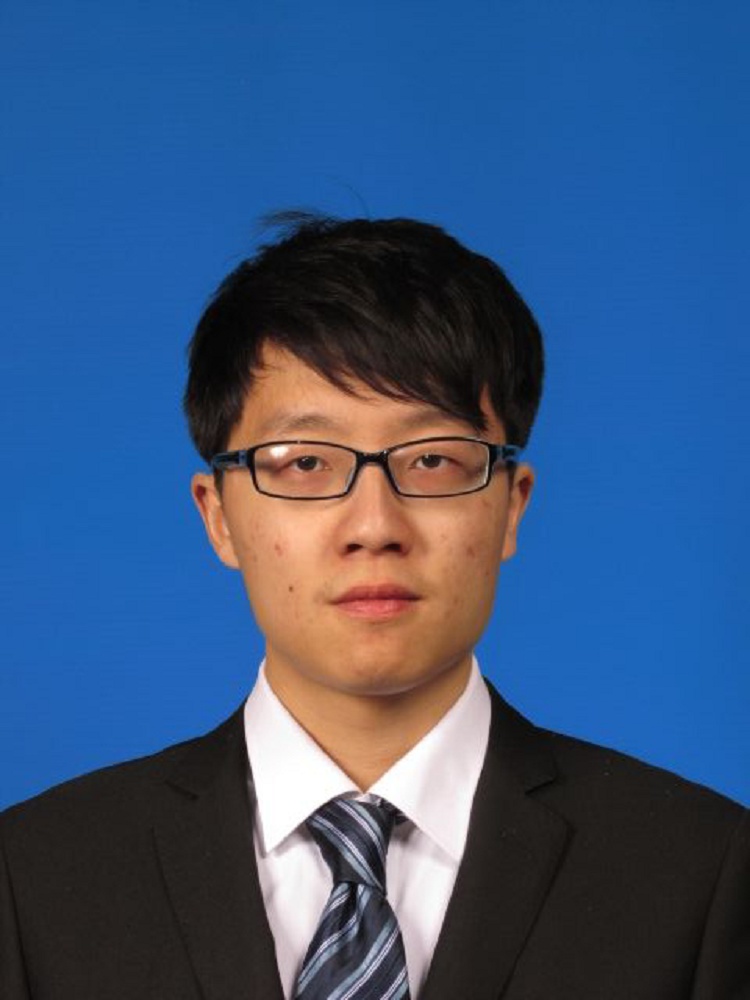}}]{Zheyan Shen}
is a Ph.D candidate in Department of Computer Science and Technology, Tsinghua University. He received his B.S. from the Department of Computer Science and Technology, Tsinghua University in 2017. His research interests include causal inference, stable prediction under selection bias and interpretability of machine learning.
\end{IEEEbiography}
\vskip -0.05in

\begin{IEEEbiography}[{\includegraphics[width=1.0in,height=1.25in,clip,keepaspectratio]{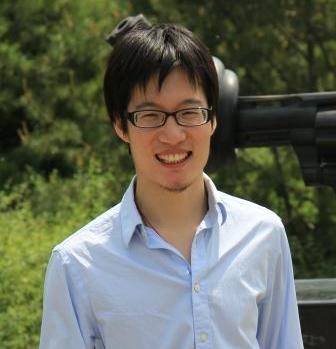}}]{Peng Cui}
is an Associate Professor with tenure in Tsinghua University. He got his PhD degree from Tsinghua University in 2010. His research interests include causally-regularized machine learning, network representation learning, and social dynamics modeling. He has published more than 100 papers in prestigious conferences and journals in data mining and multimedia. His recent research won the IEEE Multimedia Best Department Paper Award, SIGKDD 2016 Best Paper Finalist, ICDM 2015 Best Student Paper Award, SIGKDD 2014 Best Paper Finalist, IEEE ICME 2014 Best Paper Award, ACM MM12 Grand Challenge Multimodal Award, and MMM13 Best Paper Award. He is PC co-chair of CIKM2019 and MMM2020, SPC or area chair of WWW, ACM Multimedia, IJCAI, AAAI, etc., and Associate Editors of IEEE TKDE, IEEE TBD, ACM TIST, and ACM TOMM etc. He received ACM China Rising Star Award in 2015, and CCF-IEEE CS Young Scientist Award in 2018. He is now a Distinguished Member of ACM and CCF, and a Senior Member of IEEE.
\end{IEEEbiography}
\vskip -0.05in

\begin{IEEEbiography}[{\includegraphics[width=1.0in,height=1.25in,clip,keepaspectratio]{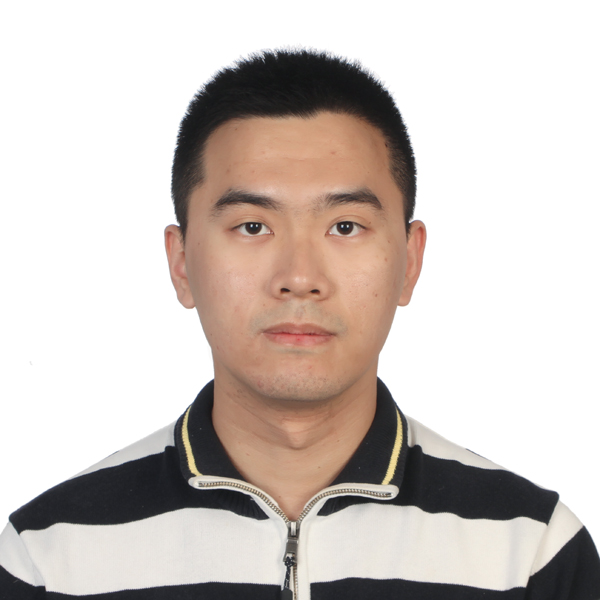}}]{Linjun Zhou}
received BE degree from the Department of Computer Science and Technology of Tsinghua University in 2016. He is a Ph.D. candidate from Lab of Media and Network, Department of Computer Science and Technology, Tsinghua University now. His research interests include few-shot learning and robust learning.
\end{IEEEbiography}

\begin{IEEEbiography}[{\includegraphics[width=1.0in,height=1.25in,clip,keepaspectratio]{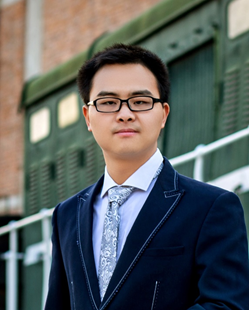}}]{Kun Kuang}, Associate Professor in the College of Computer Science and Technology, Zhejiang University. He received his Ph.D. in the Department of Computer Science and Technology at Tsinghua University in 2019. He was a visiting scholar at Stanford University. His main research interests include causal inference and causally regularized machine learning. He has published over 30 papers in major international journals and conferences, inclu ding SIGKDD, ICML, ACM MM, AAAI, IJCAI, TKDE, TKDD, Engineering, and ICDM, etc.
\end{IEEEbiography}

\begin{IEEEbiography}[{\includegraphics[width=1.0in,height=1.25in,clip,keepaspectratio]{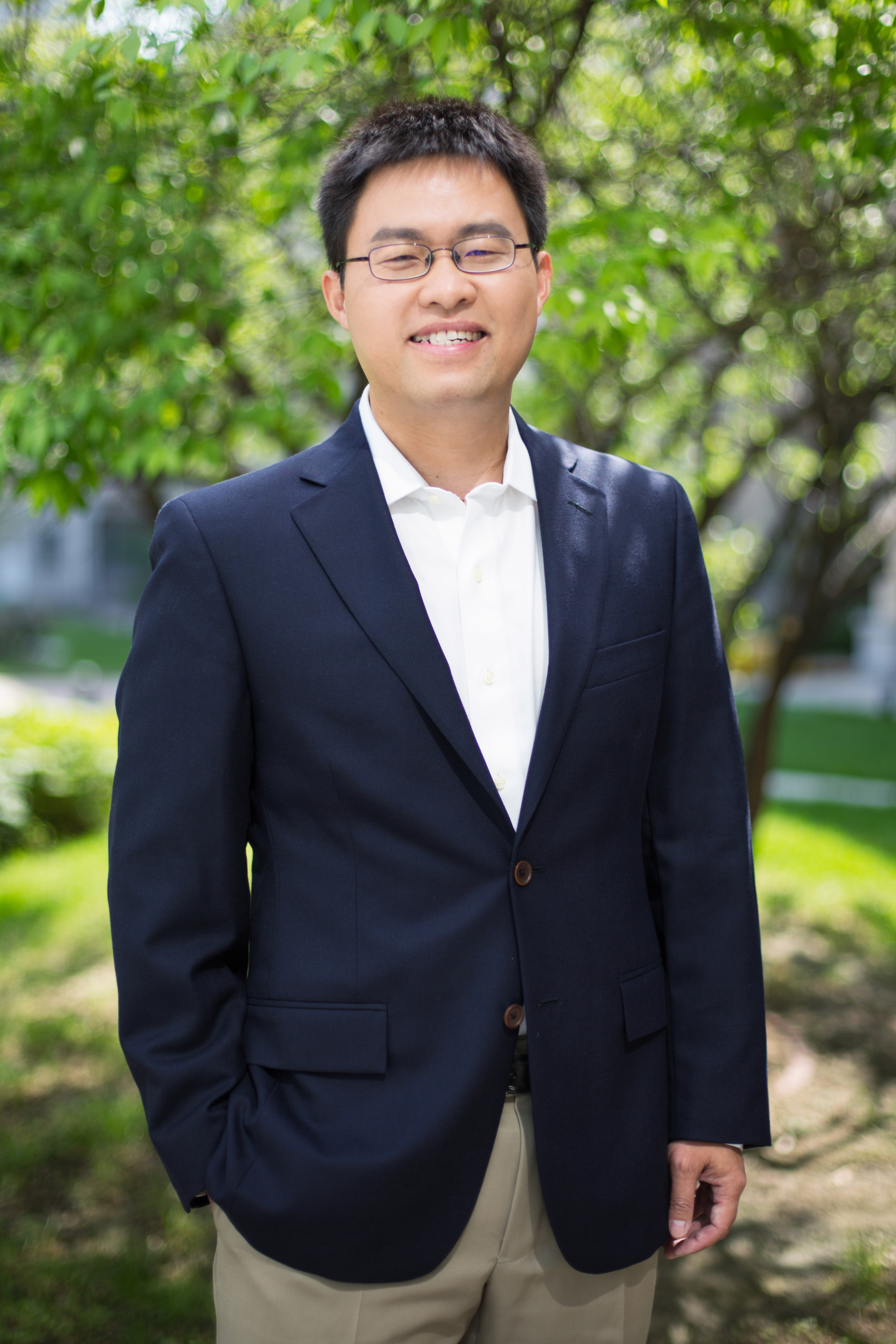}}]{Bo Li}
received a Ph.D degree in Statistics from the University of California, Berkeley, and a bachelor’s degree in Mathematics from Peking University. He is an Associate Professor at the School of Economics and Management, Tsinghua University. His research interests are business analytics and risk-sensitive Artificial Intelligence. He has published widely in academic journals across a range of fields including statistics, management science and economics.
\end{IEEEbiography}

\newpage
\appendix 
\input{data/7appendix}

%
%
%
%
%

%
%
%



\end{document}

%% file: data/1intro.tex
\IEEEraisesectionheading{\section{Introduction}\label{sec:introduction}}

%
%
%
%
\IEEEPARstart{T}{raditional} machine learning algorithms which optimize the average empirical loss often suffer from the poor generalization performance under distributional shifts induced by latent heterogeneity, unobserved confounders or selection biases in training data\cite{daume2006domain,torralba2011unbiased, shen2019stable}. 
However, in high-stake applications such as medical diagnosis\cite{kukar2003transductive}, criminal justice\cite{berk2018fairness,rudin2018optimized} and autonomous driving \cite{huval2015empirical}, it is critical for the learning algorithms to ensure the robustness against potential unseen data.
Therefore, robust learning methods have recently aroused much attention due to its favorable property of robustness guarantee\cite{ben1998robust,goodfellow2014explaining,madry2017towards}.

Instead of optimizing the empirical cost on training data, robust learning methods seek to optimize the worst-case cost over an uncertainty set and can be further separated into two main branches named adversarially and distributionally robust learning. 
In adversarially robust learning, the uncertainty set is constructed point-wisely\cite{goodfellow2014explaining,papernot2016limitations,madry2017towards,ye2018bayesian}. 
Specifically, adversarial attack is performed independently on each data point within a $L_2$ or $L_{\infty}$ norm ball around itself to maximize the loss of current classification model.
In distributionally robust learning, on the other hand, the uncertainty set is characterized on a distributional level\cite{SinhaCertifying, esfahani2018data,duchi2018learning}. 
A joint perturbation, typically measured by Wasserstein distance or $f$-divergence between distributions, is applied to the entire distribution entailed by training data.
These methods can provide robustness guarantees under distributional shifts when testing distribution is captured in the built uncertainty set. 
However, in real scenarios, to contain the possible true testing distribution, the uncertainty set is often overwhelmingly large, and results in learned models with fairly low confidence, which is also referred to as the over pessimism or the low confidence problem\cite{frogner2019incorporating,sagawa2019distributionally}.  
That is, with an overwhelmingly large set, the learner optimizes for implausible worst-case scenarios, resulting in meaningless results (e.g. the classifier assigns equal probability to all classes).
Such a problem greatly hurts the generalization ability of robust learning methods in practice.

The essential problem of the above methods lies in the construction of the uncertainty set.
To address the over pessimism of the learning algorithm, one should form a more practical uncertainty set which is likely to contain the potential distributional shifts in the future and meanwhile is as small as possible.
More specifically, in real applications, we observe that different covariates may be perturbed in a non-uniform way, which should be considered when building a practical uncertainty set. 
Taking the problem of waterbirds and landbirds classification as an example\cite{wah2011caltech}.
There exist two types of covariates where the stable covariates (e.g. representing the bird itself) preserve immutable correlations with the target across different environments, while those unstable ones (e.g. representing the background) are pretty likely to change (e.g. waterbirds on land).
Therefore, for the example above, the construction of the uncertainty set should be anisotropic which mainly focuses on the perturbation of those unstable covariates (e.g. background) to generate more practical and meaningful samples.

\begin{figure}
    \centering
    \includegraphics[width=\linewidth]{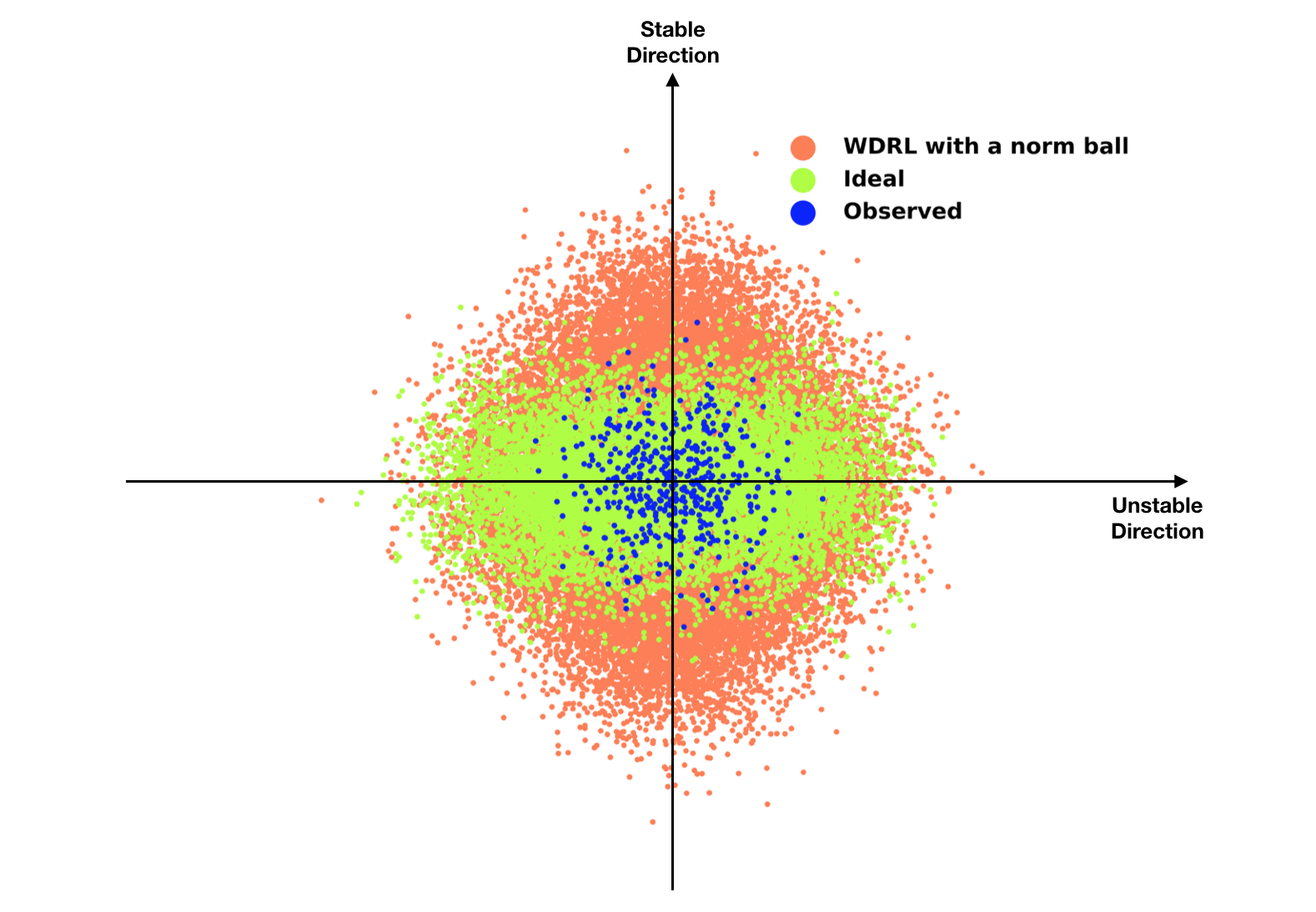}
    \caption{ Illustration for the anisotropic adversarial distribution
set, where blue points denote the observed data distribution, and
orange points denote the adversarial distribution set produce by an
isotropic Wasserstein ball, and the green one shows the ideal set
that incorporate realistic distribution shifts.}
    \label{fig:intro}
\end{figure}

Further, we illustrate the anisotropic uncertainty set in figure \ref{fig:intro}, where blue points denote the observed training distribution ($\mathcal{N}(0,I_2)$). 
And we sample data points from all distributions in the uncertainty set captured by an isotropic Wasserstein ball around the observed distribution, which are colored orange. 
We can see from figure \ref{fig:intro} that the original distribution is perturbed equally along both the stable and unstable direction. 
With the intuition above, we propose that the ideal uncertainty should be like green points, which only perturb the training distribution along unstable directions.
Following this, there are several work\cite{DBLP:journals/corr/abs-1904-06347,vaishnavi2019can} based on the adversarial attack which focus on perturbing the color or background of images to improve the adversarial robustness. 
However, these methods mainly follow a step by step routine where the segmentation is conducted first to separate the background from the foreground and cannot theoretically provide robustness guarantees under unknown distributional shifts, which greatly limits their applications on more general settings.

In this paper, we propose the Stable Adversarial Learning (SAL) algorithm to address this problem in a more principled and unified way, which leverages heterogeneous data sources to construct a more practical uncertainty set.
Specifically, we adopt the framework of Wasserstein distributionally robust learning(WDRL) and further characterize the uncertainty set to be anisotropic according to the stability of covariates across the multiple environments, which induces stronger adversarial perturbations on unstable covariates than those stable ones.  
A synergistic algorithm is designed to jointly optimize the covariates differentiating process as well as the adversarial training process of model's parameters. 
Compared with traditional robust learning techniques, the proposed method is able to provide robustness under strong distributional shifts while not hurting much confidence of the learner. 
Theoretically, we prove that our method constructs a more compact uncertainty set, which as far as we know is the first analysis of the compactness of adversarial sets in WDRL literature.
Empirically, the advantages of our SAL algorithm are demonstrated on both synthetic and real-world datasets in terms of uniformly good performance across distributional shifts. 
%

%% file: data/2related.tex
\section{Related Work}
\label{sec:related work}
In this section, we investigate several strands of related literature more thoroughly, including domain adaptation, domain generalization, stable learning and distributionally robust learning. 

Domain adaptation methods\cite{DBLP:journals/tkde/PanY10} leverage the data from target domain to assist the model training on source domain. 
Therefore the resulted model could capture the possible distributional shift in testing.
Shimodaira\cite{SHIMODAIRA2000227} proposes to assign each training data a new weight equal to the density ratio between source and target distribution, and therefore guarantee the optimality of learned model on test distribution. 
Then several techniques have been proposed to estimate the ratio more accurately, such as discriminative estimation\cite{DBLP:journals/jmlr/BickelBS09}, kernel mean matching\cite{DBLP:conf/nips/DudikSP05} and maximum entropy\cite{DBLP:conf/nips/HuangSGBS06}.
Apart from reweighting methods, deep learning based methods\cite{DBLP:conf/iccv/FernandoHST13,DBLP:conf/icml/GaninL15} learn a transformation in feature space to characterize both source and target domain.
However, the deployment of domain adaptation methods in real applications, where one can hardly access data from target domain, is quite limited. 

Compared with domain adaptation, domain generalization techniques do not require the availability of target domain data and become more and more popular these years due to its practicability. 
Different from domain adaptation, domain generalization methods propose to learn a domain-invariant classifier with multiple training domains.
Muandet et al. \cite{DBLP:conf/icml/MuandetBS13} propose a kernel-based optimization algorithm to learn an invariant latent space of data across training domains.
Through the lens of causality\cite{2015Causal,DBLP:journals/jmlr/Rojas-CarullaST18}, M. Arjovsky et al. \cite{arjovsky2019invariant} propose Invariant Risk Minimization to learn invariant representation with theoretical guarantee of the optimality of out-of-distribution generalization, which gains the most attention recently.
Also, stable learning methods \cite{kuang2018stable, shen2019stable, cui2022stable} propose to decorrelate the covariates via sample reweighting to estimate the real causal effects, which enhances the stability under distributional shifts.
However, they only deal with the covariate shift problem and do not apply to other kinds of distributional shifts (e.g. concept shifts brought by anti-causal variables).

Distributionally robust learning (DRL), from the optimization literature, proposes to optimize for the worst-case cost over an uncertainty distribution set, so as to protect the model against the potential distributional shifts in the uncertainty set, which is constrained by moment or support conditions \cite{10.1287/opre.1090.0741,bertsimas2018data-driven}, or $f$-divergence \cite{namkoong2016stochastic,sagawa2019distributionally}. 
As the uncertainty set formulated by Wasserstein ball is much more flexible, Wasserstein Distributionally Robust Learning (WDRL) has been widely studied \cite{shafieezadehabadeh2015distributionally, SinhaCertifying,esfahani2018data}. 
WDRL for logistic regression was proposed by Abadeh et al. \cite{shafieezadehabadeh2015distributionally}. 
Sinha et al. \cite{SinhaCertifying} achieved moderate levels of robustness with little computational cost relative to empirical risk minimization with  a Lagrangian penalty formulation of WDRL. 
Esfahani and Kuhn \cite{esfahani2018data} reformulated the distributionally robust optimization problems over Wasserstein balls as finite convex programs under mild assumptions. 
Although DRL offers an alternative to empirical risk minimization for robust performance under distributional perturbations, there has been work questioning its real effect in practice. 
Hu et al. \cite{hu2016does} proved that when the DRL is applied to classification tasks, the obtained classifier ends up being optimal for the observed training distribution, and the core of the proof lies in the over-flexibility of the built uncertainty set.
And Fronger et al. \cite{frogner2019incorporating} also pointed out the problem of overwhelmingly-large decision set, and they used large number of unlabeled examples to further constrain the distribution set.

%% file: data/3problem.tex
\section{Problem Setting}
As mentioned above, the uncertainty set built in WDRL is often overwhelmingly large in wild high-dimensional scenarios. 
To demonstrate this over pessimism problem of WDRL, we design a toy example in \ref{exp:toy} to show the necessity to construct a more practical uncertainty set. 
Indeed, without any prior knowledge or structural assumptions, it is quite difficult to design a practical set for robustness under distributional shifts.

Therefore, in this work, we consider a dataset $D=\left\{D^e\right\}_{e\in \mathcal{E}_{tr}}$, which is a mixture of data $D^e=\left\{(x_i^e, y_i^e)\right\}_{i=1}^{n_e}$ collected from multiple training environments $e\in \mathcal{E}_{tr}$, $x_i^e\in\mathcal{X}$ and $y_i^e\in\mathcal{Y}$ are the $i$-th data and label from environment $e$ respectively.
Specifically, each dataset $D^e$ contains examples identically and independently distributed according to some joint distribution $P_{XY}^e$ on $\mathcal{X}\times\mathcal{Y}$.
Given the observations that in real scenarios, different covariates have different extents of stability, we propose assumption \ref{assup1}. 
\begin{assumption}
\label{assup1}
	There exists a decomposition of all the covariates $X = \{S,V\}$, where $S$ represents the stable covariate set and V represents the unstable one, so that for all environments $e \in \mathcal{E}$, $\mathbb{E}[Y^e|S^e=s,V^e=v] = \mathbb{E}[Y^e|S^e=s] = \mathbb{E}[Y|S=s]$. 
\end{assumption}

Intuitively, assumption \ref{assup1} indicates that the correlation between stable covariates $S$ and the target $Y$ stays invariant across environments, which is quite similar to the invariance in \cite{DBLP:conf/kdd/KuangCAXL18, arjovsky2019invariant, DBLP:conf/aaai/KuangX0A020}. 
Moreover, assumption \ref{assup1} also demonstrates that the influence of $V$ on the target $Y$ can be wiped out as long as whole information of $S$ is accessible.
Under the assumption \ref{assup1}, the disparity among covariates revealed in the heterogeneous datasets can be leveraged for better construction of the uncertainty set. 
And based on assumption \ref{assup1}, we propose our problem:
\begin{problem}
	Given multi-environments training data $D=\{D^e\}_{e\in\mathcal{E}_{tr}}$, under assumption \ref{assup1}, the goal is to build a more practical uncertainty set for distributionally robust learning and achieve stable performance across distributional shifts with respect to low $\mathrm{Mean\_Error}$ defined as $\mathrm{Mean\_Error} = \frac{1}{|\mathcal{E}_{te}|}\sum_{e \in \mathcal{E}_{te}}\mathcal{L}^e$ and low $\mathrm{Std\_Error}$ defined as $\mathrm{Std\_Error} = \sqrt{\frac{1}{|\mathcal{E}_{te}|-1}\sum_{e\in \mathcal{E}_{te}}\left(\mathcal{L}^e - \mathrm{Mean\_Error}\right)^2}$.
\end{problem}

%% file: data/4method.tex
\section{Method}
In this work, we propose the Stable Adversarial Learning (SAL) algorithm, which leverages heterogeneous data to build a more practical uncertainty set with covariates differentiated according to their stability.

Firstly, we introduce the Wasserstein Distributionally Robust Learning (WDRL) framework which attempts to learn a model with minimal risk against the worst-case distribution in the uncertainty set characterized by Wasserstein distance: 
\begin{definition}
Let $\mathcal{Z} \subset \mathbb{R}^{m+1}$ and $\mathcal{Z} = \mathcal{X}\times\mathcal{Y}$ , given a transportation cost function $c: \mathcal{Z} \times \mathcal{Z} \rightarrow [0, \infty)$, which is nonnegative, lower semi-continuous and satisfies $c(z,z)=0$, for probability measures $P$ and $Q$ supported on $\mathcal{Z}$, the Wasserstein distance between $P$ and $Q$ is :
\begin{equation}
W_c(P, Q) = \inf\limits_{M \in \Pi(P,Q)} \mathbb{E}_{(z,z') \sim M}[c(z,z')]
\end{equation}
where $\Pi(P,Q)$ denotes the couplings with $M(A,\mathcal{Z})=P(A)$ and $M(\mathcal{Z},A)=Q(A)$ for measures $M$ on $\mathcal{Z}\times \mathcal{Z}$.
\end{definition}

Following the intuition above that the uncertainty should not be isotropic along stable and unstable directions, we propose to learn an anisotropic uncertainty set with the help of heterogeneous environments.  
The objective function of our SAL algorithm is:
\begin{small}
\begin{align}
\label{equ:obj1}
	&\min\limits_{\theta\in\Theta}\sup_{Q: W_{c_w}(Q,P_0)\leq \rho}\mathbb{E}_{X,Y\sim Q}[\ell(\theta;X,Y)]\\
\label{equ:obj2}
	\text{s.t.\ }w \in \arg&\min\limits_{w\in \mathcal{W}} \left\{ \frac{1}{|\mathcal{E}_{tr}|}\sum_{e\in\mathcal{E}_{tr}}\mathcal{L}^e(\theta) + \alpha\max\limits_{e_p,e_q\in \mathcal{E}_{tr}}\mathcal{L}^{e_p} - \mathcal{L}^{e_q}\right\}
\end{align}
\end{small}
where $P_0$ denotes the training distribution, $W_{c_w}$ denotes the Wasserstein distance with transportation cost function $c_w$ defined as
\begin{equation}
	\label{equ:cw}
	c_w(z,z')=\|w\odot(z-z')\|^2
\end{equation}
and $\mathcal{W} = \left\{w : w\in[1,+\infty)^{m+1} \ \ \text{and}\  \min(w)=1\right\}$ denotes the covariate weight space ($\min(w)$ denotes the minimal element of $w$), 
and $\mathcal{L}^e$ denotes the average loss in environment $e\in\mathcal{E}_{tr}$, $\alpha$ is a hyper-parameter to adjust the tradeoff between average performance and the stability. 

The core of our SAL is the covariate weight learning procedure in equation \ref{equ:obj2}. 
In our algorithm, the uncertainty set is built to achieve stable performance across heterogeneous multiple environments. 
Intuitively, $w$ controls the perturbation level of each covariate and formulates an anisotropic uncertainty set compared with the conventional WDRL methods.
The objective function of $w$ (equation \ref{equ:obj2}) contains two parts: the average loss in training environments as well as the maximum margin, which aims at learning such $w$ that the resulting uncertainty set leads to a learner with uniformly good performance across environments. 
Equation \ref{equ:obj1} is the objective function of model's parameters via distributionally robust learning with the learnable covariate weight $w$. 
During training, the covariate weight $w$ and model's parameters $\theta$ are iteratively optimized.

Details of the algorithm are delineated below. We first will introduce the optimization of model's parameter in section \ref{sec4-1:optimization}, then the transportation cost function learning procedure in section \ref{sec4-2:weight}. 
The pseudo-code of the whole Stable Adversarial Learning (SAL) algorithm is shown in Algorithm \ref{algo:SAL}. 

\begin{algorithm}[]
  \caption{Stable Adversarial Training}
  \label{algo:SAL}
\begin{algorithmic}
  \STATE {\bfseries Input:} Multi-environments data $D^{e_1},\ D^{e_2},\ \dots,\ D^{e_n}$, where $D^e=(X^e,Y^e)$, $e\in \mathcal{E}$
  \STATE {\bfseries Hyperparameters:} $T$, $T_{\theta}$, $T_w$, m,  $\epsilon_x$, $\epsilon_{\theta}$, $\epsilon_w$, $\alpha$
  \STATE {\bfseries Initialize:} $w = [1.0, \dots, 1.0]$
  \FOR{ $i=1$ {\bfseries to} $T$}
  
        \FOR{$j=0$ {\bfseries to} $T_{\theta}-1$}
            \STATE Initialize $\widetilde{X}_0$ as: $\widetilde{X}_0 = X$
            \FOR{$k=0$ {\bfseries to} $m-1$}
                \STATE
                \algorithmiccomment{Approximate the supreme of $s_{\lambda}(X)$ for $X^e$ from all $e\in\mathcal{E}$}
                
                \STATE $\widetilde{X}^e_{k+1} = \widetilde{X}^e_{k} + \epsilon_x\nabla_x\{\ell(\theta;\widetilde{X}^e_{k}) - \lambda c_w(\widetilde{X}^e_{k},\widetilde{X}^e_0)\}$

            \ENDFOR
            
            \STATE {\bf Update $\theta$ as:\ \ } $\theta^{j+1} = \theta^{j} - \epsilon_{\theta}\nabla_{\theta} \ell(\theta^j;(\widetilde{X}_m,Y))$
        \ENDFOR
        
        \STATE {\bf Calculate $R(\theta)$ as:\ \ }$R(\theta)=\frac{1}{|\mathcal{E}|}\sum_{e\in\mathcal{E}}\mathcal{L}^e + \alpha\left(\sup\limits_{p,q\in \mathcal{E}}\mathcal{L}^p - \mathcal{L}^q\right)$
        
        \STATE $w^0 = w^{i}$
        \FOR{$j=0$ {\bfseries to} $T_w-1$}
            \STATE {\bf Update $w$ as:\ \ }$w^{j+1}=w^j - \epsilon_w\nabla_wR(\theta)$
        \ENDFOR
        \STATE {\bfseries Update w as:\ \ }, $w^{i+1}=\mathop{Proj}_{\mathcal{W}}\left(w^{t_w}\right)$.
  \ENDFOR
\end{algorithmic}
\end{algorithm}

\subsection{Tractable Optimization}
\label{sec4-1:optimization}
In SAL algorithm, the model's parameters $\theta$ and covariate weight $w$ is optimized iteratively. In each iteration, given current $w$, the objective function for $\theta$ is:
\begin{equation}
\label{equ:sec4-obj1}
	\min\limits_{\theta\in\Theta}\sup_{Q: W_{c_w}(Q,P_0)\leq \rho}\mathbb{E}_{X,Y\sim Q}[\ell(\theta;X,Y)]
\end{equation}
The duality results in lemma \ref{theorem_refor} show that the infinite-dimensional optimization problem \ref{equ:sec4-obj1} can be reformulated as a finite-dimensional convex optimization problem \cite{esfahani2018data}. Besides, inspired by \cite{SinhaCertifying}, a Lagrangian relaxation is provided for computation efficiency.

\begin{lemma}
\label{theorem_refor}
Let $\mathcal{Z} = \mathcal{X}\times\mathcal{Y}$ and $\ell: \Theta \times \mathcal{Z} \rightarrow \mathbb{R}$ be continuous. For any distribution $Q$ and any $\rho \ge 0$, let $s_{\lambda}(\theta;(x,y)) = \sup\limits_{\xi \in \mathcal{Z}} (\ell(\theta;\xi) - \lambda c_w(\xi,(x,y)))$, $\mathcal{P}=\{Q: W_c(Q,P_0)\leq \rho\}, $we have:
	\begin{small}
	\begin{equation}
	\label{duality}
		\sup\limits_{Q \in \mathcal{P}} \mathbb{E}_Q[\ell(\theta;x,y)] = \inf\limits_{\lambda \geq 0} \{\lambda\rho + \mathbb{E}_{P_0}[s_{\lambda}]\}
	\end{equation}
	\end{small}
and for any $\lambda \geq 0$, we have:
\begin{small}
	\begin{equation}
		\label{equ:relax}
		\sup\limits_{Q}\{\mathbb{E}_Q[\ell(\theta;(x,y))]-\lambda W_{c_w}(Q,P_0)\} =\mathbb{E}_{P_0}[s_{\lambda}] 
	\end{equation}
\end{small}
\end{lemma}

The original problem \ref{equ:sec4-obj1} can firstly be reformulated as equation \ref{duality} by duality. 
However, the infimum with respect to $\lambda$ is also intractable. 
Therefore, we give up the prescribed amount $\rho$ of robustness in equation (\ref{equ:sec4-obj1}) and focus instead on the relaxed Lagrangian penalty function for efficiency in equation (\ref{equ:relax}). 
Notice that there exists only the inner supremum in $\mathbb{E}_{P_0}[s_{\lambda}(\theta;(x,y))]$ in equation (\ref{equ:relax}), which can be seen as a relaxed Lagrangian penalty function of the original objective function (\ref{equ:sec4-obj1}). 
Following lemma \ref{theorem_refor}, we derive the loss function on empirical distribution $\hat{P}_N$ as:
\begin{small}
\begin{equation}
	\hat{\mathcal{L}}(\theta)=\frac{1}{N}\sum_{i=1}^N s_{\lambda}(\theta;(x_i,y_i))
\end{equation}
\end{small}
Recall that $s_\lambda(\theta;(x,y))=\sup\limits_{\xi \in \mathcal{Z}} (\ell(\theta;\xi) - \lambda c_w(\xi,(x,y)))$, we propose to convert the minimization of $\hat{\mathcal{L}}$ over $\theta$ to a minimax procedure as done in \cite{SinhaCertifying} to approximate the supremum for $s_{\lambda}$:
\begin{small}
\begin{equation}
	\min_\theta \max_{\widetilde{X}}\mathbb{E}_{\hat{P}_N}\left[ \ell(\theta;\widetilde{X},Y) - \lambda c_w((\widetilde{X},Y),(X,Y))\right]
\end{equation}
\end{small}
Specifically, given predictor $X$, we adopt gradient ascent to obtain an approximate maximizer $\widetilde{X}$ of $s_\lambda(\theta;(X,Y))$ and optimize the model's parameter $\theta$ using $(\widetilde{X}, Y)$.
In the following parts, we simply use $\widetilde{X}$ to denote $\{\widetilde{x}\}_N$, which means the set of maximizers for training data $\{x\}_N$. 
The convergence guarantee for this optimization can be referred to \cite{SinhaCertifying}.

\subsection{Learning for transportation cost $w$}
\label{sec4-2:weight}

We introduce the learning for transportation cost function $c_w$ in this section. In supervised scenarios, perturbations are typically only added to predictor $X$ and not target $Y$. Therefore, we simplify $c_w: \mathcal{Z}\times\mathcal{Z}\rightarrow[0,+\infty)(\mathcal{Z}=\mathcal{X}\times\mathcal{Y})$ to be:
\begin{small}
\begin{align}
\label{equ:w-simplify}
	c_w(z_1,z_2) &= c_w(x_1,x_2) + \infty\times\mathbb{I}(y_1\neq y_2)\\
				 &= \|w\odot(x_1-x_2)\|_2^2 + \infty\times\mathbb{I}(y_1\neq y_2)
\end{align}
\end{small}
and omit '$y$-part' in $c_w$ as well as $w$, that is $w \in [1,+\infty)^{m}$  in the following parts. 
Intuitively, $w$ controls the strength of adversary put on each covariate. 
The higher the weight is, the weaker perturbation is put on the corresponding covariate. 
Ideally, we hope the covariate weights on stable covariates are extremely high to protect them from being perturbed and to maintain the stable correlations, while weights on unstable covariates are nearly $1$ to encourage perturbations for breaking the harmful spurious correlations. 
With the goal towards uniformly good performance across environments, we come up with the objective function $R(\theta(w))$ for learning $w$ as:
\begin{small}
\begin{equation}
	R(\theta(w))=\frac{1}{|\mathcal{E}_{tr}|}\sum_{e\in\mathcal{E}_{tr}}\mathcal{L}^e(\theta(w)) + \alpha\max\limits_{e_p,e_q\in \mathcal{E}_{tr}}\left(\mathcal{L}^{e_p} - \mathcal{L}^{e_q}\right)
\end{equation}
\end{small}
where $\alpha$ is the hyper-parameter. $R(\theta(w))$ contains two parts: the first is the average loss in multiple training environments; the second reflects the max margin among environments, which reflects the stability of $\theta(w)$, since it is easy to prove that $\max\limits_{e_p,e_q\in \mathcal{E}_{tr}}\mathcal{L}^{e_p}(\theta(w)) - \mathcal{L}^{e_q}(\theta(w))=0$ if and only if the errors among all training environments are same. 
Here $\alpha$ is used to adjust the tradeoff between average performance and stability.

Given current $\theta^t$, we can update $w$ as:
\begin{small}
\begin{equation}
	w^{t+1} = Proj_{\mathcal{W}}\left(w^t - \epsilon_w \frac{\partial R(\theta^t)}{\partial w}\right)
\end{equation}
\end{small}
where $Proj_{\mathcal{W}}$ means projecting onto the space $\mathcal{W}$. 
And the remaining work is how to calculate the gradient $\partial R(\theta(w))/\partial w$, which we will introduce in detail in following section \ref{sec:calculation_gradient}.

\subsubsection{Calculation of $\partial R(\theta(w))/\partial w$}
\label{sec:calculation_gradient}
In order to optimize $w$, $\partial R(\theta(w))/\partial w$ can be approximated as following.
\begin{small} 
\begin{equation}
	\frac{\partial R(\theta(w))}{\partial w} = \frac{\partial R}{\partial \theta}\frac{\partial \theta}{\partial X_A}\frac{\partial X_A}{\partial w}
\end{equation}
\end{small}
Note that the first term $\partial R/ \partial \theta$ can be calculated easily. The second term can be approximated during the gradient descent process of $\theta$ as :
\begin{small}
\begin{align}
	\theta^{t+1} &= \theta^t - \epsilon_{\theta}\nabla_{\theta}\hat{\mathcal{L}}(\theta^t;\widetilde{X},Y)\\
	\frac{\partial \theta^{t+1}}{\partial \widetilde{X}} &= \frac{\partial \theta^{t}}{\partial \widetilde{X}} - \epsilon\frac{\nabla_{\theta}\hat{\mathcal{L}}(\theta^t;\widetilde{X},Y)}{\partial \widetilde{X}}\\
	\frac{\partial \theta}{\partial \widetilde{X}} &\approx -\epsilon\sum_t \frac{\nabla_{\theta}\hat{\mathcal{L}}(\theta^t;\widetilde{X},Y)}{\partial \widetilde{X}}
\end{align}
\end{small}
where $\frac{\nabla_{\theta}\hat{\mathcal{L}}(\theta^t;\widetilde{X},Y)}{\partial \widetilde{X}}$ can be calculated during the training process. The third term $\partial \widetilde{X} / \partial w$ can be approximated during the adversarial learning process of $\widetilde{X}$ as:
\begin{small}
\begin{align}
	\widetilde{X}^{t+1} &= \widetilde{X}^t + \epsilon_x\nabla_{\widetilde{X}^t}\left\{\ell(\theta;\widetilde{X}^t,Y) - \lambda c_w(\widetilde{X}^t,X)\right\}\\
	\frac{\partial \widetilde{X}^{t+1}}{\partial w} &= \frac{\partial \widetilde{X}^{t}}{\partial w} - 2\epsilon_x\lambda \mathrm{Diag}\left(\widetilde{X}^t - X\right)\\
	\frac{\partial \widetilde{X}}{\partial w} &\approx -2\epsilon_x\lambda \sum_t \mathrm{Diag}(\widetilde{X}^t - X)
\end{align}
\end{small}
which can be accumulated during the adversarial training process. 

\subsubsection{Approximation precision}
We approximate the $\partial \theta / \partial \widetilde{X}$ and $\partial \widetilde{X} / \partial w$ during the gradient descent and ascent process, where we use the average gradient as the approximate value.
To better quantify the precision of our approximation, we tested the reliability of our approximation empirically.
Since the gradient represents the direction to which the function declines fastest, we compare the $\Delta R$ after updating by our $\partial R / \partial w$ with that after randomly selected directions with the same step size.
Note that the $\Delta R$ brought by the accurate gradient is largest among any other directions.
Therefore, the higher possibility that our $\Delta R$ is larger than randomly picked direction, the more accurate our approximation is.
We perform random experiments for 1000 runs, and the approximation of our SAL outperforms 99.4\% of them, which validates the high precision of our approximation.

%% file: data/5theory.tex
\section{Theoretical Analysis}
Here we first provide the robustness guarantee for our method, and then we analyze the rationality of our uncertainty set, which also demonstrates the uncertainty set built in our SAL is more practical.
And we finally derive the generalization bounds for our method.
\subsection{Robustness Guarantee}
Recall that the original objective of this work is to optimize for the worst-case error in a distribution set, which is given as $\min_{\theta\in\Theta}\sup_{Q:W_{c_w}(Q,P_0)\leq \rho}\mathbb{E}[\ell(\theta)]$.
However, for tractable optimization in section \ref{sec4-1:optimization}, we have to give up the prescribed amount $\rho$ of the distributional robustness and focus on the relaxed Lagrangian penalty function:
\begin{small}
\begin{equation}
\label{equ:new}
		\sup\limits_{Q}\left\{\mathbb{E}_Q[\ell(\theta;(x,y))]-\lambda W_{c_w}(Q,P_0)\right\}
\end{equation}
\end{small}	
Note that in equation \ref{equ:new}, we do not impose any constraints (e.g., within a Wasserstein ball) on the $Q$, which we optimize the equation \ref{equ:new} with respect to.
Then a natural question is, can the relaxed Lagrangian reformulation, which we actually optimize, provide some kind of robustness guarantee? 
Or it is just an approximation? 
In this subsection, we derive the robustness guarantee for the relaxed Lagrangian reformulation to answer this question.

\begin{theorem}[Robustness Guarantee for Relaxed Lagrangian Reformulation]
\label{theorem:robuseness}
For fixed $\lambda\geq 0$, define the transportation map $T_\lambda(\theta;z_0)=\arg\max_{\xi\in\mathcal{Z}}\ell(\theta;\xi)-\lambda c_w(\xi,z_0)$, and the empirical maximizer of the Lagrangian reformulation (equation \ref{equ:new}) is given as:
\begin{small}
\begin{equation}
	P_n^* = \arg\max_Q \left\{\mathbb{E}_Q[\ell(\theta;(x,y))]-\lambda W_{c_w}(Q,\hat{P}_n)\right\}
\end{equation}
\end{small}
Then we denote the Wasserstein distance between the worst-case distribution $P_n^*$ and the training distribution $\hat{P}_n$ as $\hat{\rho}_n = W_{c_w}(P_n^*, \hat{P}_n)$, 
we have:
\begin{small}
\begin{equation}
\begin{aligned}
	\sup\limits_{P:W_{c_w}(P,\hat{P}_n)\leq \hat{\rho}_n}\mathbb{E}_{P}[\ell(\theta;Z)]&=\mathbb{E}_{\hat{P}_n}[s_\lambda(\theta;Z)]+\lambda \hat{\rho}_n\\
	 &= \mathbb{E}_{P_n^*}[\ell(\theta;Z)]+\lambda\hat{\rho}_n
\end{aligned}
\end{equation}
\end{small}
\end{theorem}
\begin{proof}
	By choosing $\hat{\rho}_n$ as $\rho$ in Lemma \ref{theorem_refor}, it is easy to prove under the strong duality.
\end{proof}
Theorem \ref{theorem:robuseness} justifies that our relaxed Lagrangian reformulation in optimization can exactly guarantee the distributional robustness inside a $\hat{\rho}_n$-radius ball, that is, given $\lambda$, our algorithm will find a distribution $P_n^*$, whose distance from the original $\hat{P}_n$ is $\hat{\rho}_n$, and we can guarantee that the learned $P_n^*$ is exactly the worst-case distribution in the $\hat{\rho}_n$-radius ball centered at $\hat{P}_n$. 
The only difference from the direct optimization is that, we cannot guarantee the robustness for a pre-given quantity $\rho$, while we use the Lagrangian parameter $\lambda$ as a qualitative factor to control how much robustness to protect.

\subsection{Compactness of the Adversarial Set}
Then we analyze the rationality of our method in theorem \ref{theorem:sparse}, where our major theoretical contribution lies on. 
As far as we know, it is the first analysis of the compactness of adversary sets in WDRL literature.

\begin{assumption}
\label{assump:theoretical}
	Given $\rho>0$, $\exists Q_0 \in \mathcal{P}_0$ that satisfies: 
	
	$\mathrm{(1)}$\ $\forall \epsilon>0$, $\left|\inf\limits_{M \in \Pi(P_0,Q_0)}\mathbb{E}_{(z_1,z_2 \sim M)}\left[c(z_1,z_2)\right]\right| \leq \epsilon$, we refer to the couple minimizing the expectation as $M_0$.
	
	$\mathrm{(2)}$\ $\mathbb{E}_{M\in \Pi(P_0,Q_0)-M_0}\left[c(z_1,z_2)\right]\geq \rho$, where $\Pi(P_0,Q_0)-M_0$ means excluding $M_0$ from $\Pi(P_0,Q_0)$.
	
	$\mathrm{(3)}$\ $Q_{0\#S}\neq P_{0\#S}$, where $S=\{i: w^{(i)}>1\}$ and $w^{(i)}$ denotes the $i$th element of $w$ and $P_{\#S}$ denotes the marginal distribution on dimensions $S$.
\end{assumption} 

\begin{assumption}
\label{assump:appendix}
	Given $\rho\geq 0$ and $c_w$, there exists distribution $V$ supported on $\mathcal{Z}_{\#U}$ that
	\begin{small}
	\begin{equation}
		W_{c_w}(V, P_{0\#U})=\rho
	\end{equation}
	\end{small}
\end{assumption}

Assumption \ref{assump:theoretical} describes the boundary property of the original uncertainty set $\mathcal{P}_0 = \{Q:W_c(Q,P_o)\leq \rho\}$, which assumes that there exists at least one distribution on the boundary whose marginal distribution on $S$ is not the same as the center distribution $P_0$'s and is easily satisfied. 
And Assumption \ref{assump:appendix} assumes that there exists at least one marginal distribution $V$ whose distance from the original marginal distribution is $\rho$, and is easily satisfied.
Based on these assumptions, we come up with the following theorem.

\begin{theorem}[Compactness]
\label{theorem:sparse}
Under Assumption \ref{assump:theoretical}, assume the transportation cost function in Wasserstein distance takes form of $c(x_1,x_2)=\|x_1-x_2\|_1$ or $c(x_1,x_2)=\|x_1-x_2\|_2^2$. Then, given observed distribution $P_0$ supported on $\mathcal{Z}$ and $\rho \geq 0$, for the adversary set $\mathcal{P} = \{Q:W_{c_w}(Q,P_0)\leq \rho\}$ and the original $\mathcal{P}_0 = \{Q:W_{c}(Q,P_0)\leq \rho\}$, given $c_w$ where $\min(w^{(1)},\dots, w^{(m)})=1$ and $\max(w^{(1)},\dots, w^{(m)})>1$, we have $\mathcal{P} \subset \mathcal{P}_0$. Furthermore, under Assumption \ref{assump:appendix}, for the set $U=\{i | w^{(i)}=1\}$, $\exists Q_0 \in \mathcal{P}$ that satisfies $W_{c_w}(P_{0\#U},Q_{0\#U})=\rho$.
\end{theorem}
Theorem \ref{theorem:sparse} proves that the constructed uncertainty set of our method is smaller than the original. 
Intuitively, in adversarial learning paradigm, if stable covariates are perturbed, the target should also change correspondingly to maintain the underlying relationship.
However, we have no access to the target value corresponding to the perturbed stable covariates in practice, so optimizing under an isotropic uncertainty set (e.g. $P_0$) which contains perturbations on both stable and unstable covariates would generally lower the confidence of the learner and produce meaningless results. 
Therefore, from this point of view, by adding high weights on stable covariates in the cost function, we may construct a more reasonable and practical uncertainty set in which the ineffective perturbations are avoided.

Further, we theoretically analysis the property of learned covariate weights $w$ in linear regression, including the optimal point of equation \ref{equ:obj2} and the reason why our method can to some extent mitigate the low confidence problem compared with the original WDRL.
To begin with, we make further assumptions on the given multiple environments data.
\begin{assumption}[Data Heterogeneity]
\label{assumption: heterogeneous}
	Under Assumption \ref{assup1}, we further assume that $\exists \delta_S \geq 0, \delta_V > 0$, such that:\\ (1) $\forall e\in\mathcal{E}$, $|\min_\theta \mathcal{L}^e(\theta)-\min_{\theta_S}\mathcal{L}^e(\theta_S)|\leq \delta_S$\\
	(2) $\forall$ linear model $f_\theta(X)=\theta_S^TS+\theta_V^TV$ with $\theta_V>0$, $\exists e_i,e_i\in\mathcal{E}_{tr}$ such that $\mathcal{L}^{e_i}(\theta)-\mathcal{L}^{e_j}(\theta)>\delta_V$.
	where $\theta_S$ denotes the linear parameters on stable covariates and $\theta_V$ on unstable covariates.
\end{assumption}
Actually, Assumption \ref{assumption: heterogeneous} assumes that (1) the predicting performance with stable features or unstable features will not differ much; (2) using unstable features for prediction will hurt the model's stability across different environments, since $\mathbb{E}^e[Y|V]$ may change greatly.
\begin{theorem}[Optimal $\theta^*(w^*)$]
\label{theorem:optimal}
	Under Assumption \ref{assumption: heterogeneous}, for $\alpha > \frac{\delta_S}{\delta_V}$, the optimal point $\theta^*(w^*)$ of equation \ref{equ:obj2} satisfies that $\theta_V^*=0$ and $w^*_V$.
	Further, choosing $c(z_1,z_2)=\|z_1-z_2\|_2$, with $\rho\rightarrow\infty, \rho^2/w^*_S\rightarrow 0$ and $w^*_V=1$, the minimizer $\theta'$ of equation \ref{equ:obj1} will approach to $\theta^*$.
\end{theorem}
\begin{proof}
	 It is easy to prove the parameters of unstable features in $\theta^*$ is 0 under Assumption \ref{assumption: heterogeneous}.We move on to the property of $w^*$. 
	 For $c_w=\|w\odot(z_1-z_2)\|_2$, the equation \ref{equ:obj1} can be reformulated to (following \cite{esfahani2018data})
	 \begin{small}
	 \begin{equation}
	 \label{equ:ridge-like}
	 	\theta' = \arg\min_\theta \frac{1}{N}\sum_{i=1}^N\ell_i+\rho\sqrt{(-\theta,1)^T\text{Diag}^{-1}(w)(-\theta,1)}
	 \end{equation}
	 \end{small}
	 Then with $\rho\rightarrow\infty, \rho^2/w^*_S\rightarrow 0, \rho^2w^*_V\rightarrow \infty$, it is easy to prove that $\theta'_V\rightarrow 0$ and $\theta'_S = \arg\min_{\theta_S}\mathcal{L}$.
\end{proof}
In Theorem \ref{theorem:optimal}, we analyze the properties of the optimal points of our method, which verifies that the learned covariate weights will greatly restrict the perturbations on stable features ($w^*_S\rightarrow \infty$) to mitigate the over-pessimism problem.
Although the scenario is simple, we can also get inspirations why the original WDRL faces the low confidence problem. 
From the reformulation in equation \ref{equ:ridge-like}, we see that WDRL regulates the predictor with $\|(-\theta,1)\|_2$ (by letting $w=1$) and the strength of regularization is controlled by the radius $\rho$ of the ball.
As $\rho$ grows to contain more potential testing distributions, WDRL puts much more penalty on the parameters of both stable features and unstable features, which lowers both $\theta_S$ and $\theta_V$ until they are both 0, making in the model refuse to make predictions and only output 0, that is the origin of low confidence or over-pessimism.
While in our proposed method, we use the learned covariate weights $w$ to prevent the parameters $\theta_S$ of stable features from being affected, and such desired weight can be learned via equation \ref{equ:obj2} as shown in Theorem \ref{theorem:optimal}.

\subsection{Generalization Bounds}
First, we provide the robustness guarantee in theorem \ref{theorem:bound} with the help of lemma \ref{theorem_refor} and Rademacher complexity\cite{bartlett2002rademacher}. 
\begin{theorem}[Generalization Bounds]
\label{theorem:bound}
Let $\Theta=R^m,\ x\in \mathcal{X},\ y\in\mathcal{Y}$.
Assume $|\ell(\theta;z)|$ is bounded by $T_{\ell} \geq 0$ for all $\theta\in \Theta,\ z=(x,y)\in\mathcal{X}\times\mathcal{Y}$.
Let $F:\mathcal{X}\rightarrow\mathcal{Y}$ be a class of prediction functions, then for $\theta\in \Theta,\ \rho \geq 0,\ \lambda \geq 0$, with probability at least $1-\delta$, for $P \in \{P:W_{c_w}(P,P_0)\leq \rho\}$, we have:
\begin{small}
\begin{equation}
    \begin{aligned}
        \sup\limits_{P}\mathbb{E}_P\left[\ell(\theta;Z)\right] \leq \lambda\rho + &\mathbb{E}_{\hat{P}_n}\left[s_{\lambda}(\theta;Z)\right] + \mathcal{R}_n(\widetilde{\ell}\circ F) \\ 
        & + kT_{\ell}\sqrt{\ln(1/\delta)/n}
    \end{aligned}
\end{equation}
\end{small}
where $\widetilde{\ell}\circ F=\{(x,y)\mapsto \ell(f(x),y)-\ell(0,y):f\in F\}$ and $\mathcal{R}_n$ denotes the Rademacher complexity\cite{bartlett2002rademacher} and $k$ is a numerical constant no less than 0. 
\end{theorem}
\begin{proof}
From lemma \ref{theorem_refor}, for all $\lambda \geq 0,\ \rho \geq 0$, we have
\begin{equation}
\label{equ:proof1-1}
	\sup\limits_{P:W_{c_w}(P,P_0)}\mathbb{E}_{P}[\ell(\theta;X,Y)]\leq \lambda\rho + \mathbb{E}_{P_0}[s_{\lambda}(\theta;X,Y)]
\end{equation}
Applying the standard results on Rademacher complexity\cite{bartlett2002rademacher}, with probability at least $1-\delta$, we have:
\begin{small}
\begin{align}
	\mathbb{E}_{P_0}[s_{\lambda}] \leq \mathbb{E}_{\hat{P}_n}[s_{\lambda}] + \mathcal{R}_n(\widetilde{l}\circ F) +kT_{\ell}\sqrt{\frac{\ln(1/\delta)}{n}}
\end{align}
\end{small}
then combing with equation \ref{equ:proof1-1}, the result follows.
\end{proof}
Since the Rademacher complexity $\mathcal{R}_n$ also requires the expectation over sample distribution, we further derive the bound of the Rademacher complexity in theorem \ref{theorem:bound} which only depends on empirical data points.
We introduce the definition of $\epsilon$-cover and $\epsilon$-covering number as follows, which can be used to measure the size of  continuous sets.
\begin{definition}[$\epsilon$-cover]
	$\mathcal{C}\subset \mathcal{U}$ is an $\epsilon$-cover of a functional class $\mathcal{G}\subset\mathcal{U}$ if and only if for all $g \in\mathcal{G}$, there exists some $h \in \mathcal{C}$ such that $d_n(g,h)\leq \epsilon$, where $d_n(\cdot,\cdot)$ is function distance metric defined with respect to a tuple of data points $(z_1, \dots, z_n)\in\mathbb{R}^d$ as:
	\begin{small}
	\begin{equation}
	\label{equ:functional distance metric}
		d_n(g,h) = \sqrt{\frac{1}{n}\sum_{i=1}^n(g(z_i)-h(z_i))^2}
	\end{equation} 
	\end{small}
\end{definition}
\begin{definition}[$\epsilon$-covering number]
	The $\epsilon$-covering number of a function class $\mathcal{G}$ is defined as:
	\begin{equation}
		N(\mathcal{G}, \epsilon, d_n(\cdot, \cdot)) = \inf \{|\mathcal{C}|: \mathcal{C}\ is\ an\ \epsilon\mathrm{-cover}\ of\  \mathcal{G}\}
	\end{equation}
	where $d_n(\cdot,\cdot)$ denotes the function distance metric as equation \ref{equ:functional distance metric}. 
\end{definition}
Then we derive the bound of Rademacher complexity $\mathcal{R}_n$ with respect to the $\epsilon$-covering number.
\begin{theorem}[$\hat{\mathcal{R}}_n$]
\label{theorem:hat}
	For the Rademacher complexity in theorem \ref{theorem:bound}, for function set $\mathcal{G}$ and assume that $\forall g\in \mathcal{G}$, $g:\mathcal{Z}\rightarrow \mathbb{R}$ is a function and is bounded by $T_\ell \geq 0$, with probability at least $1-\delta$, we have:
	\begin{small}
	\begin{equation}
		\mathcal{R}_n(\mathcal{G}) \leq \hat{\mathcal{R}}_n(\mathcal{G}) + 2T_\ell\sqrt{\log 1/\delta/2n}
	\end{equation}
	\end{small}
\end{theorem}
\begin{proof}
	Easy to prove with bounded difference inequality. 
\end{proof}
Finally, we would like to derive the bound for $\hat{\mathcal{R}}_n$ with $\epsilon$-covering number.
\begin{theorem}(Bound of $\hat{\mathcal{R}}_n$)
\label{theorem: bound of hat}
	For function class $\mathcal{G}$ containing functions $G:\mathcal{Z}\rightarrow \mathbb{R}$, we have:
	\begin{small}
	\begin{equation}
		\hat{\mathcal{R}}_n(\mathcal{G}) \leq \inf_{\epsilon\geq 0}\left\{4\epsilon + 12\int_{\epsilon}^{\sup_{G\in\mathcal{G}}\sqrt{\hat{\mathbb{E}}[G^2]}}\sqrt{\log N(\mathcal{G},\tau,d_n(\cdot,\cdot))/n}d\tau\right\} 
	\end{equation}
	\end{small}
	Specifically, assume that $\forall G\in\mathcal{G}:\mathcal{Z}\rightarrow \mathbb{R}$, $|G|$ is bounded by $T_\ell \geq 0$, we have:
	\begin{small}
	\begin{equation}
		\hat{\mathcal{R}}_n(\mathcal{G}) \leq \inf_{\epsilon\geq 0}\left\{4\epsilon + 12\int_{\epsilon}^{T_\ell}\sqrt{\frac{\log N(\mathcal{G},\tau,d_n(\cdot,\cdot))}{n}}d\tau\right\} 
	\end{equation}
	\end{small}
\end{theorem}
\begin{proof}
	Let $\tau_0=\sup_{G\in\mathcal{G}}\sqrt{\hat{\mathbb{E}}[G^2]}$ and for any $j \in \mathbb{Z}_+$ let $\tau_j=2^{-j}\tau_0$. For each $j$, let $\mathcal{C}_j$ be a $\tau_j$-cover of $\mathcal{G}$ with respect to $d_n(\cdot,\cdot)$. For each $G \in \mathcal{G}$ and $j$, pick an $\hat{G_j}\in\mathcal{C}_j$ such that $\hat{G_j}$ is an $\alpha_j$ approximation of $G$. Then for $N \in \mathbb{Z}_+$, $G$ can be expressed as $G = G - \hat{G}_N + \sum_{i=1}^N(\hat{G}_N - \hat{G}_{N-1})$ where $\hat{G}_0 = 0$. Then for any $N$, we have:
	\begin{footnotesize}
		\begin{align}
		\hat{\mathcal{R}}_n(\mathcal{G}) &= \frac{1}{n}\mathbb{E}_\sigma [\sup_{G\in\mathcal{G}}\sum_{i=1}^n\sigma_i(G(x_i)-\hat{G}_N(x_i) + \sum_{j=1}^N(\hat{G}_j(x_i)-\hat{G}_{j-1}(x_i)))]\\
		&\leq \tau_N + \sum_{j=1}^N\frac{1}{n}\mathbb{E}_\sigma \left[\sup_{G\in\mathcal{G}}\sum_{i=1}^n\sigma_i(\hat{G}_j(x_i)-\hat{G}_{j-1}(x_i))\right]
	\end{align}
	\end{footnotesize}
	\begin{small}
	\begin{align}
		\text{Note that\quad}d_n(\hat{G}_j - \hat{G}_{j-1})^2 &= d_n(\hat{G}_j - G + G - \hat{G}_{j-1})^2\\
		&\leq (\tau_j + \tau_{j-1})^2 = 9\tau_j^2
	\end{align}
	\end{small}
	Then apply Massart's finite class lemma to function classes $\{f-f': f\in\mathcal{C}_j, f'\in\mathcal{C}_{j-1}\}$(for each $j$), we have for any $N$ that,
	\begin{small}
	\begin{align}
		\hat{R}_n(\mathcal{G}) \leq \tau_N + 12\int_{\alpha_{N+1}}^{\alpha_0}\sqrt{\frac{\log N(\mathcal{G}, \tau, d_n(\cdot,\cdot))}{n}}d\tau
	\end{align}
	\end{small}
	Then for any $\epsilon$, choose $N=\sup\{j : \alpha_j>2\epsilon\}$. We have $\alpha_N \leq 4\epsilon$ and
	\begin{small}
	\begin{equation}
		\hat{\mathcal{R}}_n(\mathcal{G}) \leq 4\epsilon + 12\int_{\epsilon}^{\sup_{G\in\mathcal{G}}\sqrt{\hat{\mathbb{E}}[G^2]}}\sqrt{\frac{\log N(\mathcal{G},\tau,d_n(\cdot,\cdot))}{n}}d\tau
	\end{equation}
	\end{small}
	Since $\epsilon$ is arbitrarily chosen, we take an infimum over $\epsilon$.
\end{proof}
\begin{remark}
	Merging Theorem \ref{theorem:bound}, \ref{theorem:hat}, \ref{theorem: bound of hat} together, we obtain the final bound as:
	\begin{small}
		\begin{equation}
		\begin{aligned}
			&\sup_P\mathbb{E}_P[\ell(\theta;Z)]\leq \lambda \rho + \mathbb{E}_{\hat{P}}[s_\lambda(\theta;Z)] +kT_\ell\sqrt{\frac{\log(1/\delta)}{n}}+\\ 
			&\inf_{\epsilon\geq 0}\left\{4\epsilon + 12\int_{\epsilon}^{\sup_{G\in\mathcal{G}}\sqrt{\hat{\mathbb{E}}[G^2]}}\sqrt{\frac{\log N(\mathcal{G},\tau,d_n(\cdot,\cdot))}{n}}d\tau\right\}
		\end{aligned}
		\end{equation}	
	\end{small}

\end{remark}

%% file: data/6experiments.tex
\section{Experiments}
In this section, we validate the effectiveness of our method on simulation data and real-world data.

{\bf Baselines\ \ \ } We compare our proposed SAL with the following methods. 
\begin{itemize}
\small
	\item Empirical Risk Minimization(ERM):
	    \begin{equation}
	        \min\limits_{\theta}\mathbb{E}_{P_0}\left[\ell(\theta;X,Y)\right]
	    \end{equation}
	\item Wasserstein Distributionally Robust Learning(WDRL):
	    \begin{equation}
	        \min\limits_{\theta}\sup\limits_{Q \in W(Q,P_0)\leq \rho}\mathbb{E}_{Q}\left[\ell(\theta;X,Y)\right]
	    \end{equation}
	\item Invariant Risk Minimization(IRM\cite{arjovsky2019invariant}):
	    \begin{equation}
	        \min\limits_{\theta}\sum_{e\in\mathcal{E}}\mathcal{L}^e+\lambda\|\nabla_{w|w=1.0}\mathcal{L}^e(w\cdot\theta)\|^2
	    \end{equation}
\end{itemize}
For completeness, we also compare with LASSO \cite{lasso},
and Ridge regression \cite{hoerl1970ridge}.

For ERM and WDRL, we simply pool the multiple environments data for training. 
For fairness, we search the hyper-parameter $\lambda$ in $\{0.01, 0.1, \dots, 1e0, 1e1,\dots,1e4\}$ for IRM and the hyper-parameter $\rho$ in $\{1, 5, 10, 20,50,80,100\}$ for WDRL. 
And we search the hyper-parameters $\lambda$ for LASSO and Ridge in $\{1e-3, 1e-2, \dots, 1e-1, \dots, 1e1\}$. 
The best hyper-parameter is selected according to the validation set, which is sampled i.i.d from the training environments.

\textbf{Kinds of Distributional Shifts\ } To demonstrate the superiority of our methods, we design two typical kinds of distributional shifts, including \emph{selection bias}\cite{kuang2018stable, cui2022stable} and \emph{anti-causal effects}\cite{arjovsky2019invariant}.
In our simulation data, we introduce \emph{strong distributional shifts}, where the spurious correlation between training and testing data varies a lot.

{\bf Evaluation Metrics\ } We use $\mathrm{Mean\_Error}$ defined as $\mathrm{Mean\_Error} = \frac{1}{|\mathcal{E}_{te}|}\sum_{e \in \mathcal{E}_{te}}\mathcal{L}^e$ and $\mathrm{Std\_Error}$ defined as $\mathrm{Std\_Error} = \sqrt{\frac{1}{|\mathcal{E}_{te}|-1}\sum_{e\in \mathcal{E}_{te}}\left(\mathcal{L}^e - \mathrm{Mean\_Error}\right)^2}$
which are the mean and standard deviation error across testing environments $e\in\mathcal{E}_{te}$. 

{\bf Imbalanced Mixture} In our experiments, we perform a non-uniform sampling among different environments in training set which follows the natural phenomena that empirical data follow a power-law distribution. It is widely accepted that only a few environments/subgroups are common and the rest majority are rare\cite{2018Causally,sagawa2019distributionally,2020An}.

\subsection{Simulation Data}
Firstly, we design one toy example to demonstrate the over pessimism problem of conventional WDRL. 
Then, we design two mechanisms to simulate the varying correlations of unstable covariates across environments, named by selection bias and anti-causal effect. 

\subsubsection{Toy Example}
\label{exp:toy}

\begin{figure*}[ht]
\vskip -0.1in
\subfigure[Testing performance for each environment.]{
\label{img:toy-radius2}
\includegraphics[width=0.32\linewidth]{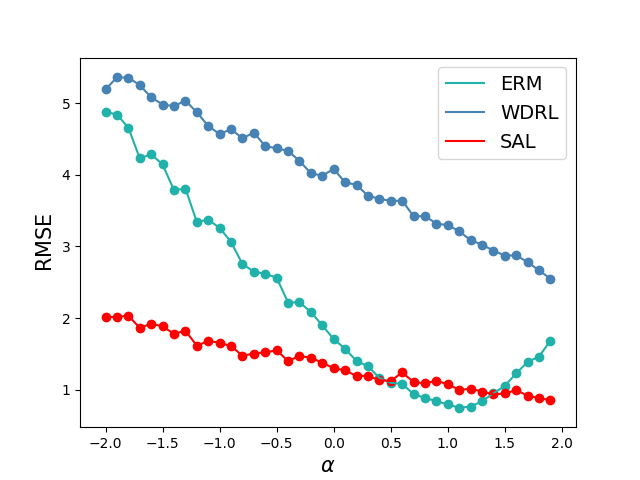}}
\subfigure[Testing performance with respect to radius]{
\label{img:toy-radius}
\includegraphics[width=0.32\linewidth]{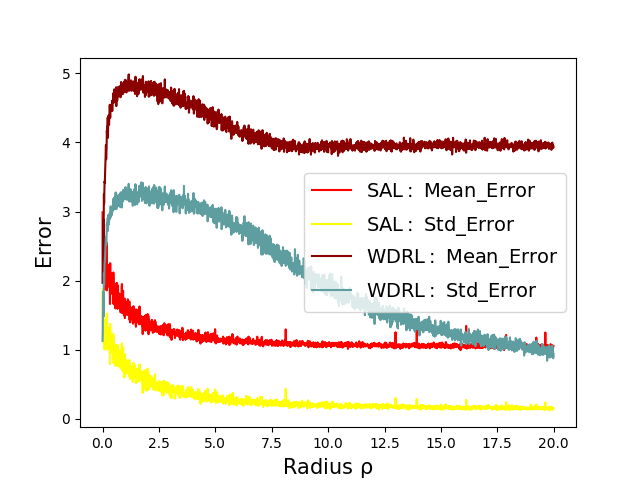}}
\subfigure[The learned coefficients of $S$ and $V$ w.r.t. radius]{
\label{img:toy-estimate}
\includegraphics[width=0.32\linewidth]{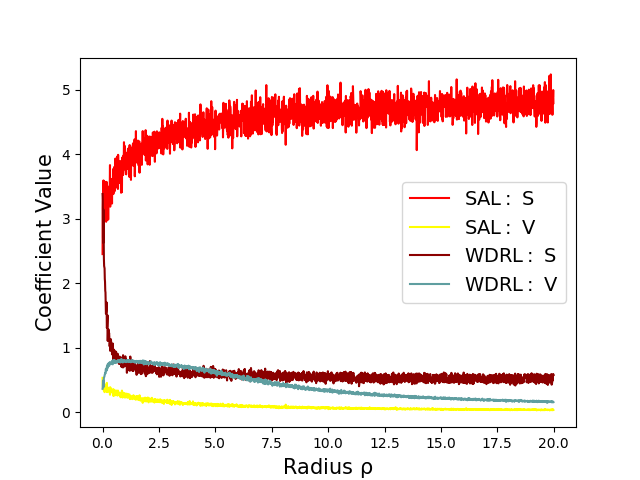}}

\caption{Results of the toy example. The left figure shows the testing performance in different environments under fixed radius, where $\mathrm{RMSE}$ is root mean square error for the prediction. The middle and right denotes the prediction error and the learned coefficients of WDRL and SAL w.r.t. radius.}

\vskip -0.1in
\end{figure*}
\label{sec:toy}

\begin{figure*}[ht]
\vskip -0.1in
\subfigure[Visualization of the stable feature $S$ and  unstable feature $V$.]{
\label{img:toy-vis1}
\includegraphics[width=0.32\linewidth]{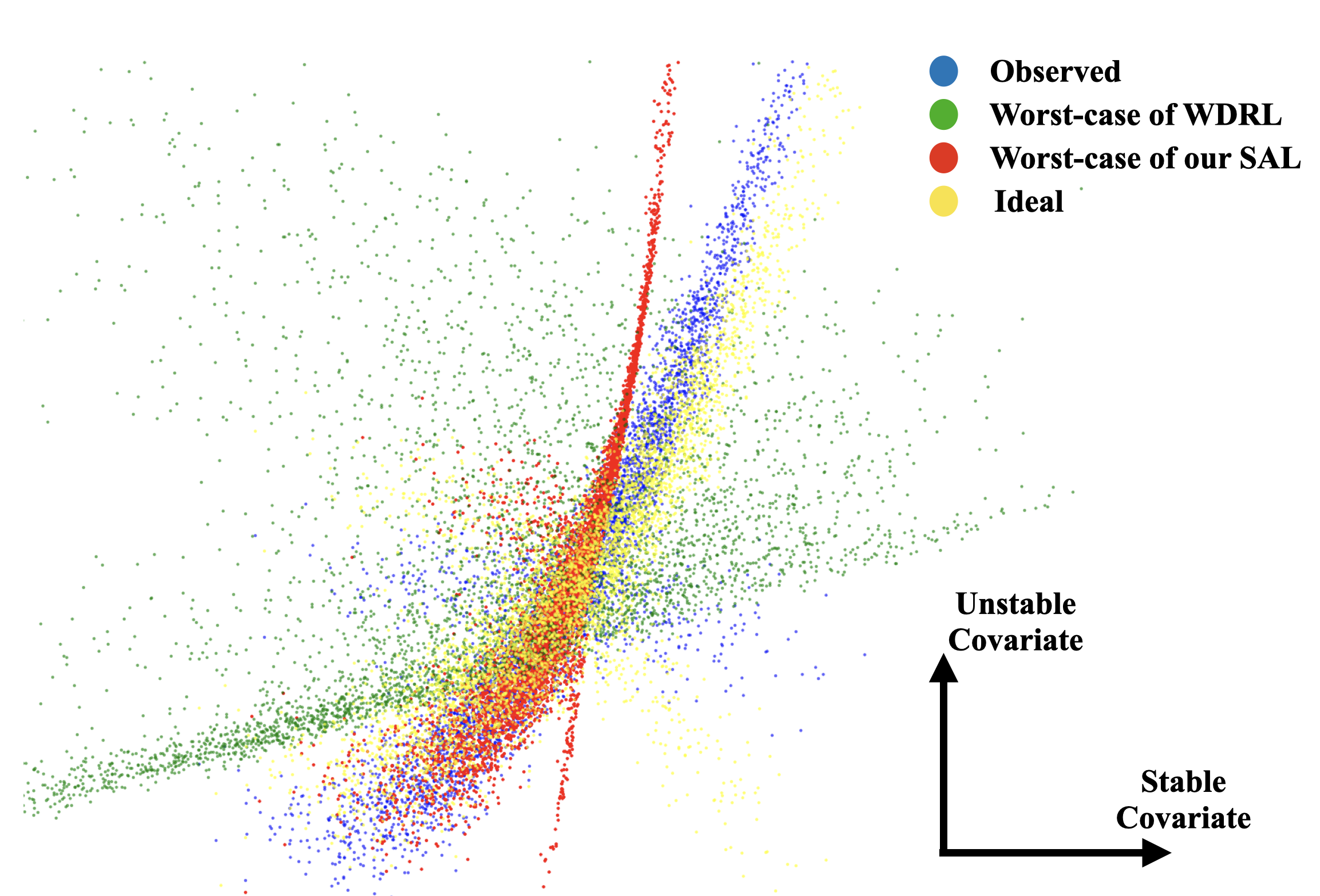}}
\subfigure[Visualization of the stable feature $S$ and  target $Y$.]{
\label{img:toy-vis2}
\includegraphics[width=0.32\linewidth]{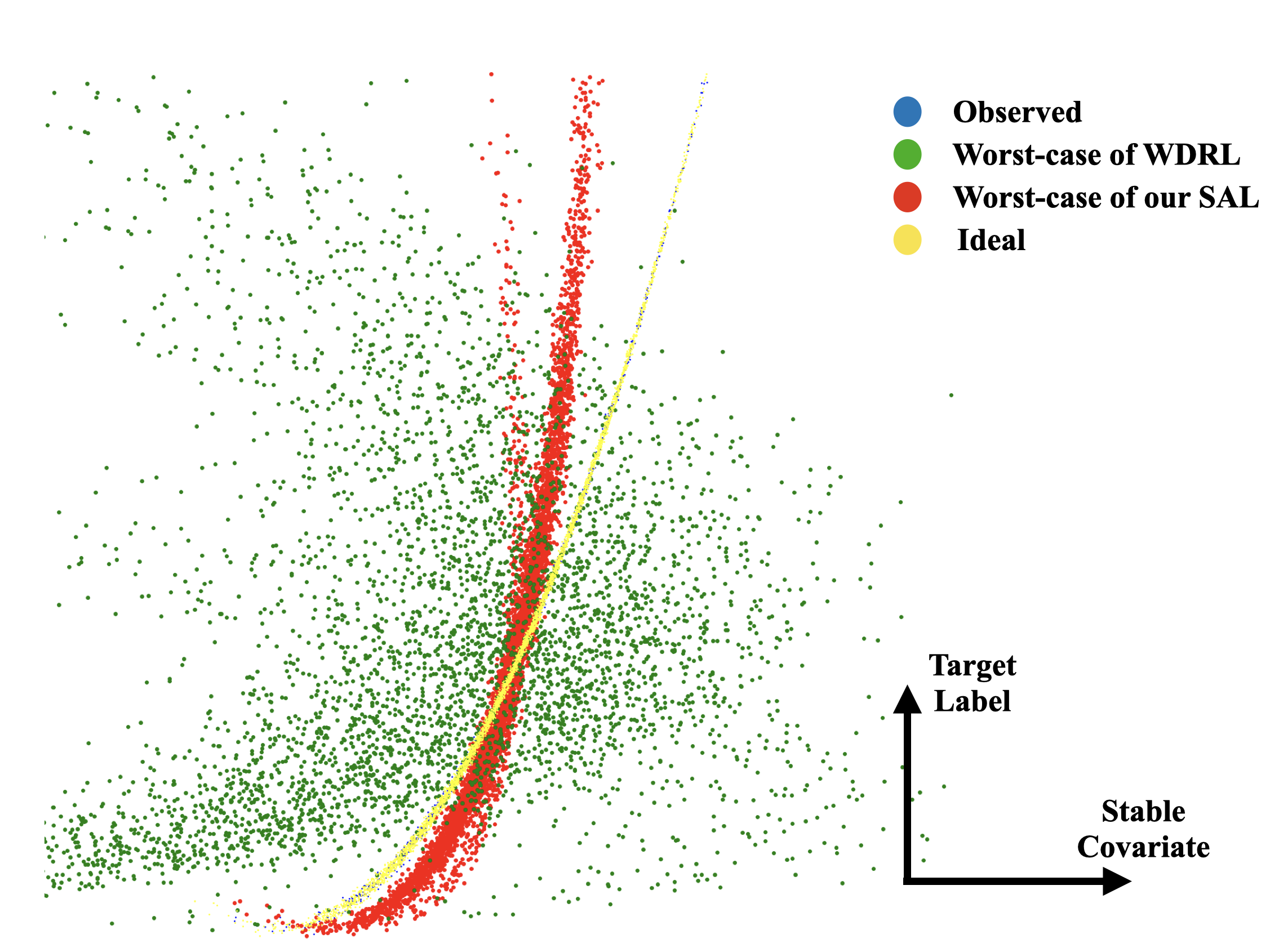}}
\subfigure[Visualization of the unstable feature $V$ and  target $Y$.]{
\label{img:toy-vis3}
\includegraphics[width=0.32\linewidth]{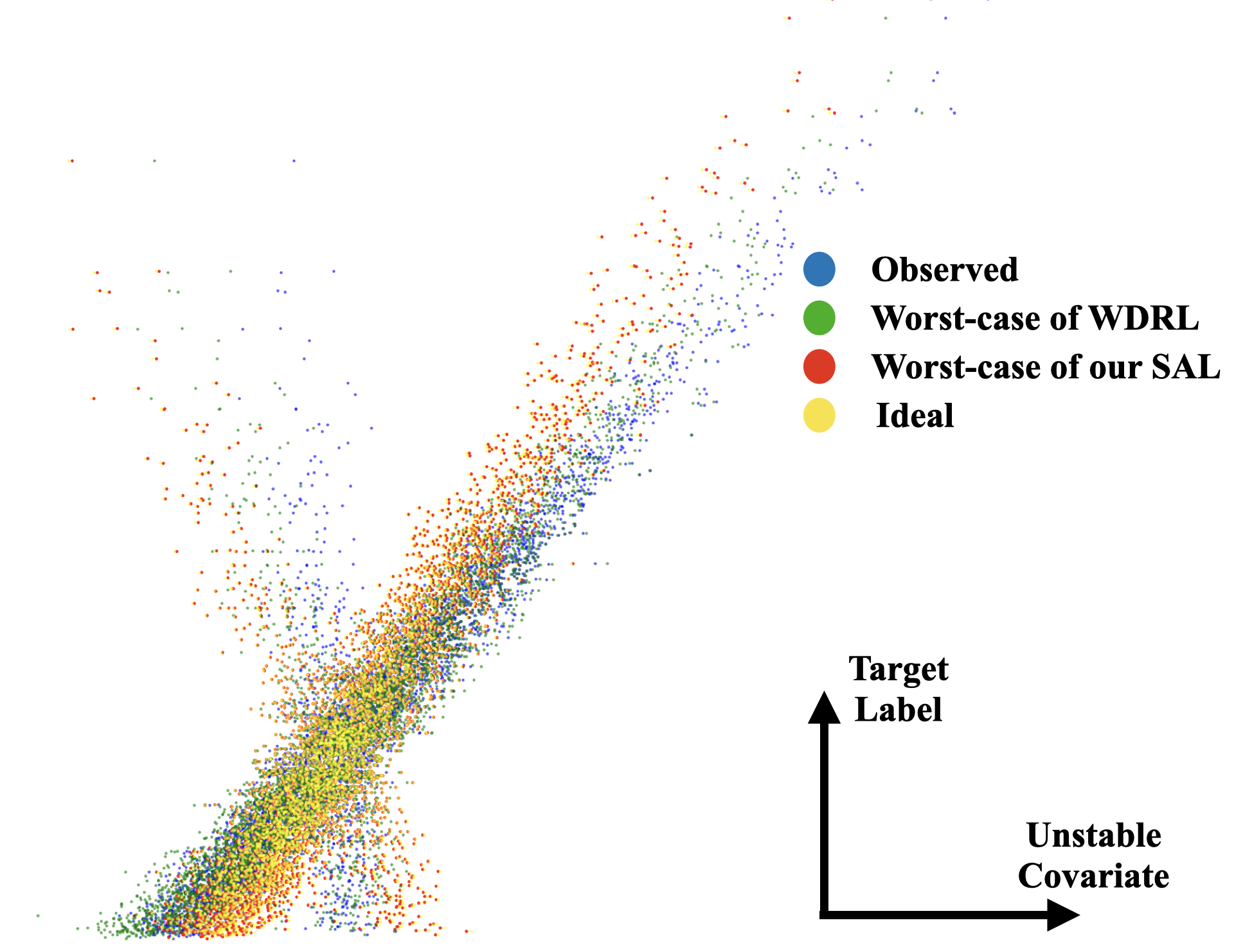}}

\caption{Visualization of the toy example. We plot the observed data points, as well as the learned worst-case distribution of WDRL, our SAL and the ideal case. The first subfigure visualizes the stable covariate $S$ and the unstable  $V$, and the second one shows $S$ and $Y$, and the third one shows $V$ and $Y$.}
\label{fig:vis}
\vskip -0.1in
\end{figure*}
\label{sec:toy}

In this setting, the goal is to predict $y \in \mathcal{R}$ from $x \in \mathcal{R}^d$, and we use $\ell(\theta;(x,y)) = |y - \theta^Tx|$ as the loss function. 
We take $d=2$ and generate $X = [S,V]^T$, where $S\stackrel{iid}{\sim} \mathcal{N}(0,0.5)$. We then generate $Y$ and $V$ as following:
\begin{small}
\begin{align}
	Y = 5*S + S^2 + \epsilon_1,\ V = \alpha Y + \epsilon_2 
\end{align}
\end{small}
where $\epsilon_1 \stackrel{iid}{\sim} \mathcal{N}(0,0.1)$ and $\epsilon_2 \stackrel{iid}{\sim} \mathcal{N}(0,1.0)$.
In this experiment, the effect of $S$ on $Y$ stays invariant, but the correlation between  $V$ and $Y$, i.e. the parameter $\alpha$, varies across environments.
In training, we generate 180 data points with $\alpha=1$ for environment 1 and 20 data points with $\alpha=-0.1$ for environment 2. 
We compared methods for linear regression across testing environments with $\alpha \in \{-2.0,-1.5,\dots,1.5,2.0\}$. 

We first set the radius for WDRL and SAL to be 20.0, and the results are shown in Figure \ref{img:toy-radius2}. 
We find the ERM induces high estimation error as it puts high regression coefficient on $V$. 
Therefore, it performs poor in terms of prediction error under distribution shifts. 
While WDRL achieves more robust performances than ERM across environments, the prediction error is much higher than the others.
Our method SAL achieves not only the smallest prediction error, but also the most robust performance across environments.

Furthermore, we train SAL and WDRL for linear regression with a varying radius $\rho \in \{0.0, 0.01,\dots,20.0\}$. 
From the results shown in Figure \ref{img:toy-radius}, we can see that, with the radius growing larger, the robustness of WDRL becomes better, but meanwhile, its performance maintains poor in terms of high $Mean\_Error$ and much worse than ERM ($\rho=0$). 
This further verifies the limitation of WDRL with respect to the overwhelmingly-large adversary distribution set. 
In contrast, SAL achieves not only better prediction performance but also better robustness across environments.  
The plausible reason for the performance difference between WDRL and SAL can be explained by Figure \ref{img:toy-estimate}.
As the radius $\rho$ grows larger, WDRL tends to conservatively estimate small coefficients for both $S$ and $V$ so that the model can produce robust prediction performances over the overwhelmingly-large uncertainty set. 
Comparatively, as our SAL provides a mechanism to differentiate covariates and focus on the robustness optimization over unstable ones, the learned coefficient of unstable covariate $V$ is gradually decreased to improve robustness, while the coefficient of stable covariate $S$ does not change much to guarantee high prediction accuracy.

To better demonstrate the superiority of our proposed SAL, we further visualize the learned worst-case distribution of WDRL, out SAL compared with the observed data points in Figure \ref{fig:vis}.
From Figure \ref{img:toy-vis1}, we can see that WDRL (green points) perturbs the observed data greatly along both the stable and unstable direction, while the learned perturbations of our SAL (red points) mainly focus on the unstable direction, which is similar to the ideal case.
To better understand why the original distribution set of WDRL is undesirable, we draw Figure \ref{img:toy-vis2}, which shows the relationship between the stable covariate $S$ and the target $Y$.
It show that WDRL (green points) greatly affects such stable relationship, while the proposed SAL does not hurt much, which is analogous to the ideal case.
From Figure \ref{img:toy-vis3}, we can see that our proposed SAL greatly perturbs the relationship between the unstable feature $V$ and target $Y$.

\vskip -0.3in
\subsubsection{Selection Bias}
In this setting, the correlations between unstable covariates and the target are perturbed through selection bias mechanism. 
We assume $X = [S,V]^T \in \mathcal{R}^p$ and $S = [S_1, S_2, \dots, S_{n_s}]^T \in \mathcal{R}^{n_s}$ is independent from $V = [V_1, V_2, \dots, V_{n_v}]\in \mathcal{R}^{n_v}$ while the covariates in $S$ are dependent with each other. 
According to assumption \ref{assup1}, we assume $Y = f(S) + \epsilon$ and $P(Y|S)$ remains invariant across environments while $P(Y|V)$ can arbitrarily change. 

Therefore, we generate training data points with the help of auxiliary variables $Z \in \mathcal{R}^d$ as following:
\begin{small}
\begin{align}
&Z_1, \dots, Z_d \stackrel{iid}{\sim} \mathcal{N}(0,1.0),\ V_1, \dots, V_{n_v} \stackrel{iid}{\sim} \mathcal{N}(0,1.0) \\
&S_i = 0.8*Z_i + 0.2 * Z_{i+1} \ \ \ \ \ for \ \ i = 1, \dots, n_s
\end{align}
\end{small}
To induce model misspecification, we generate $Y$ as:
\begin{equation}
Y = f(S) + \epsilon = \theta_s*S^T + \beta*S_1S_2S_3+\epsilon
\end{equation}
where $\theta_s = [\frac{1}{3},-\frac{2}{3}, 1, -\frac{1}{3}, \frac{2}{3}, -1 , \dots] \in \mathcal{R}^{n_s}$, and $\epsilon \sim \mathcal{N}(0, 0.3)$. 
As we assume that $P(Y|S)$ remains unchanged while $P(Y|V)$ can vary across environments, we design a data selection mechanism to induce this kind of distribution shifts.
For simplicity, we select data points according to a certain variable set $V_b \subset V$:
\begin{small}
\begin{align}
&\hat{P} = \Pi_{v_i \in V_b}|r|^{-5*|f(s) - sign(r)*v_i|} \ \mu \sim Uni(0,1 ) \\
&M(r;(x,y)) =
\begin{cases}
1, \ \ \ \ \ &\text{$\mu \leq \hat{P}$ } \\
0, \ \ \ \ \ &\text{otherwise}
\end{cases} 
\end{align}  
\end{small}
where $|r| > 1$ and $V_b \in \mathcal{R}^{n_b}$.
Given a certain $r$, a data point $(x,y)$ is selected if and only if $M(r;(x,y))=1$ (i.e. if $r>0$, a data point whose $v_i$ is close to its $y$ is more probably to be selected.)
Intuitively, $r$ eventually controls the strengths and direction of the spurious correlation between $V_b$ and $Y$(i.e. if $r>0$, a data point whose $V_b$ is close to its $y$ is more probably to be selected.).
The larger value of $|r|$ means the stronger spurious correlation between $V_b$ and $Y$, and $r \ge 0$ means positive correlation and vice versa. 
Therefore, here we use $r$ to define different environments.
In training, we generate $n$ data points, where $\kappa n$ points from environment $e_1$ with a predefined $r$ and $(1-\kappa)n$ points from $e_2$ with $r=-1.1$. In testing, we generate data points for 10 environments with $r \in [-3,-2,-1.7,\dots,1.7,2,3]$. $\beta$ is set to 1.0.

\begin{table*}[htbp]
	\centering
	\caption{Results in selection bias simulation experiments of different methods with varying selection bias $r$, ratio $\kappa$, sample size $n$ and unstable covariates' dimension $n_b$ of training data, and each result is averaged over ten times runs.}
	\label{tab:sim2}
	\vskip 0.05in
	
	\resizebox{0.8\textwidth}{47mm}{
	\begin{tabular}{|l|c|c|c|c|c|c|}
		\hline
		\multicolumn{7}{|c|}{\textbf{Scenario 1: varying selection bias rate $r$\quad($n=2000,p=10,\kappa=0.95,n_b=1$)}}\\
		\hline
		$r$&\multicolumn{2}{|c|}{$r=1.5$}&\multicolumn{2}{|c|}{$r=1.7$}&\multicolumn{2}{|c|}{$r=2.0$}\\
		\hline
		Methods &  $\mathrm{Mean\_Error}$ & $\mathrm{Std\_Error}$ &$\mathrm{Mean\_Error}$ & $\mathrm{Std\_Error}$ &  $\mathrm{Mean\_Error}$ & $\mathrm{Std\_Error}$  \\
		\hline 
		ERM & 0.484 & 0.058 & 0.561 & 0.124  & 0.572 & 0.140  \\
		LASSO&0.482 & 0.046 & 0.561 & 0.124  & 0.572 & 0.140\\
		Ridge& 0.483&0.045 & 0.560 & 0.125 & 0.572 & 0.140\\
		WDRL& 0.482 & 0.044& 0.550 & 0.114  & 0.532 & 0.112 \\
		IRM & 0.475 &\bf 0.014 & 0.464 &\bf 0.015 & 0.477 &\bf 0.015 \\
		\hline
		SAL &\bf 0.450 & 0.019 &\bf  0.449 &\bf  0.015  &\bf 0.452 & 0.017 \\
		\hline
		\multicolumn{7}{|c|}{\textbf{Scenario 2: varying ratio $\kappa$ and sample size $n$\quad($p=10,r = 1.7,n_b=1$)}}\\
		\hline
		$\kappa,n$&\multicolumn{2}{|c|}{$\kappa=0.90, n=500$}&\multicolumn{2}{|c|}{$\kappa=0.90, n=1000$}&\multicolumn{2}{|c|}{$\kappa=0.975, n=4000$}\\
		\hline
		Methods &  $\mathrm{Mean\_Error}$ & $\mathrm{Std\_Error}$ &$\mathrm{Mean\_Error}$ & $\mathrm{Std\_Error}$ &  $\mathrm{Mean\_Error}$ & $\mathrm{Std\_Error}$  \\
		\hline %
		ERM & 0.580 & 0.103 & 0.562 & 0.113  & 0.555 & 0.110 \\ 
		LASSO&0.562 & 0.110 & 0.514 & 0.078  & 0.555 & 0.122\\
		Ridge&0.561 & 0.107 & 0.517 & 0.080  & 0.555 & 0.121\\
		WDRL & 0.563 & 0.101& 0.527 & 0.083 & 0.536 & 0.108 \\
		IRM & 0.460 &\bf 0.014 & 0.464 &\bf 0.015 & 0.459	&\bf 0.014\\
		\hline
		SAL &\bf 0.454 & 0.015 & \bf 0.451 & \bf 0.015 & \bf 0.448 & \bf 0.014\\
		\hline

		\multicolumn{7}{|c|}{\textbf{Scenario 3: varying ratio $\kappa$ and sample size $n$($p=10$, $r=2.0$, $n_b=3$)}}\\
		\hline
		$\kappa, n$&\multicolumn{2}{|c|}{$\kappa=0.9,n=1000$}&\multicolumn{2}{|c|}{$\kappa=0.95,n=2000$}&\multicolumn{2}{|c|}{$\kappa=0.975,n=4000$}\\
		\hline
		Methods &  $\mathrm{Mean\_Error}$ & $\mathrm{Std\_Error}$ &$\mathrm{Mean\_Error}$ & $\mathrm{Std\_Error}$ &  $\mathrm{Mean\_Error}$ & $\mathrm{Std\_Error}$  \\
		\hline 
		ERM & 0.440 & 0.069 & 0.466 & 0.105  & 0.489 & 0.133  \\
		LASSO&0.433 & 0.059 & 0.460 & 0.097  & 0.482 & 0.124\\
		Ridge&0.434 & 0.061 & 0.457 & 0.095 & 0.481 & 0.124\\
		WDRL& 0.433 & 0.058 & 0.459 & 0.095  & 0.481 & 0.122 \\
		IRM & 0.458 &\bf 0.007 & 0.458 &\bf 0.008 & 0.458 &\bf 0.008 \\
		\hline
		SAL &\bf 0.415 & 0.019 &\bf  0.411 &  0.015  &\bf 0.411 & 0.016 \\
		\hline
	\end{tabular}
	}
\end{table*}

We compare our SAL with ERM, LASSO, Ridge, IRM and WDRL for Linear Regression. We conduct extensive experiments with different settings on $r$, $n$, $n_b$ and $\kappa$. 
In each setting, we carry out the procedure 10 times and report the average results. 
The results are shown in Table \ref{tab:sim2}.  

From the results, we have the following observations and analysis:
{\bf ERM}(as well as \textbf{LASSO \& Ridge}) suffers from the distributional shifts in testing and yields poor performance in most of the settings.
Compared with ERM, the other three robust learning methods achieve better average performance due to the consideration of robustness during the training process.
When the distributional shift becomes serious as $r$ grows, {\bf WDRL} suffers from the overwhelmingly-large distribution set and performs poorly in terms of prediction error, which is consistent with our analysis.
{\bf IRM} sacrifices the average performance for the stability across environments, which might owe to its harsh requirements on the diversity of different training environments. 
Compared with other robust learning baselines, our {\bf SAL} achieves nearly perfect performance with respect to average performance and stability, which reflects the effectiveness of assigning different weights to covariates for constructing the uncertainty set.

\subsubsection{Illustration of the Confidence Problem}
As mentioned above, WDRL is faced with the low confidence problem, which is also called the over-pessimism problem. 
We conduct a classification experiment to directly show the confidence problem of WDRL as well as the superiority of our SAL. 
We make a slight modification to the selection bias setting and turn it into a classification problem. 
Specifically, we modify the generation of $Y$ as:
\begin{equation}
    Y = \mathrm{sign}(\theta_s*S^T + \beta*S_1S_2S_3+\epsilon)
\end{equation}
where $\text{sign}(x)=1_{x\geq 0}$.
In this experiment, we set $n=2000$, $\kappa=0.95$, $p=10$, $n_b=1$ and compare the SAL with WDRL under radius of $\{1e-2, 1e-1, 1e0, 1e1\}$. 
The confidence of a binary classifier $f_\theta(.)$ is defined as the maximal prediction possibility assigned to classes:
\begin{small}
\begin{equation}
    \mathrm{Conf} = \mathbb{E}[\max(f_\theta(x), 1-f_\theta(x))]
\end{equation}
\end{small}
We report the accuracy and confidence of SAL and WDRL in Table \ref{tab:confidence}.
As the radius of the uncertainty set increasing, the confidence of a WDRL classifier decreases sharply to 0.5, which means that the binary classifier cannot make a decision and it just randomly guess the answer. 
\begin{small}
 \begin{table}[htbp]
	\small
	\centering
	\caption{Results of the classification problem under selection bias setting. $\mathrm{Acc}$ denotes the average accuracy and $\mathrm{Conf}$ the confidence.}
	\label{tab:confidence}
	\vskip 0.05in
	
	\resizebox{\columnwidth}{8mm}{
	\begin{tabular}{|l|c|c|c|c|c|c|c|c|}
		\hline
		\multicolumn{9}{|c|}{\textbf{Classification under selection bias($n=2000,p=10,\kappa=0.95,n_b=1$)}}\\
		\hline
		$\mathrm{Radius}$&\multicolumn{2}{|c|}{1e-2}&\multicolumn{2}{|c|}{1e-1}&\multicolumn{2}{|c|}{1e0}&\multicolumn{2}{|c|}{1e1}\\
		\hline
		Methods &  $\mathrm{Acc}$ & $\mathrm{Conf}$ & $\mathrm{Acc}$ & $\mathrm{Conf}$ &  $\mathrm{Acc}$ & $\mathrm{Conf}$ & $\mathrm{Acc}$ & $\mathrm{Conf}$  \\
		\hline 
		WDRL& 0.765 & 0.702& 0.581 & 0.585  & 0.377 & 0.529 & 0.361 & 0.504\\
		SAL &\bf 0.799 &\bf 0.759 &\bf  0.812 &\bf  0.785  &\bf 0.818 &\bf 0.811&\bf 0.824 & \bf 0.817\\
		\hline
	\end{tabular}
	}
\end{table}
\end{small}

\subsubsection{Anti-causal Effect}
Inspired by \cite{arjovsky2019invariant}, in this setting, we introduce the spurious correlation by using anti-causal relationship from the target $Y$ to the unstable covariates $V$.
Assume $X=[S,V]^T \in \mathcal{R}^m$ and $S = [S_1, \dots, S_{n_s}]^T\in \mathcal{R}^{n_s}$, $V=[V_1, \dots, V_{n_v}]^T\in \mathcal{R}^{n_v}$, and the data generation process is as following:
\begin{small}
\begin{align}
	S &\sim \sum_{i=1}^k z_k \mathcal{N}(\mu_i,I), Y = \theta_s^TS + \beta S_1S_2S_3+\mathcal{N}(0,0.3)\\
	V &= \theta_v Y + \mathcal{N}(0,\sigma(\mu_i)^2)
\end{align}  
\end{small}
where $\sum_{i=1}^k z_i = 1\ \&\  z_i >= 0$ is the mixture weight of $k$ Gaussian components, $\sigma(\mu_i)$ means the Gaussian noise added to $V$ depends on which component stable covariates $S$ belong to and $\theta_v \in \mathcal{R}^{n_v}$. 
Intuitively, in different Gaussian components, the corresponding correlations between $V$ and $Y$ are varying due to the different value of $\sigma(\mu_i)$. 
The larger the $\sigma(\mu_i)$ is, the weaker correlation between $V$ and $Y$ is. 

We use the mixture weight $Z=[z_1,\dots,z_k]^T$ to define different environments, where different mixture weights represent different overall strength of the effect $Y$ on $V$.
In this experiment, we set $\beta=0.1$ and build 10 environments with varying $\sigma$ and the dimension of $S,V$, the first three for training and the last seven for testing. 
Specifically, we set $\beta=0.1$, $\mu_1=[0,0,0,1,1]^T,\mu_2 = [0,0,0,1,-1]^T,\mu_2=[0,0,0,-1,1]^T,\mu_4=\mu_5=\dots=\mu_{10}=[0,0,0,-1,-1]^T$, $\sigma(\mu_1)=0.2, \sigma(\mu_2)=0.5,\sigma(\mu_3)=1.0$ and $[\sigma(\mu_4), \sigma(\mu_5),\dots,\sigma(\mu_{10})]=[3.0,5.0,\dots,15.0]$. 
$\theta_s, \theta_v$ are randomly sampled from $\mathcal{N}(1,I_5)$ and $\mathcal{N}(0,0.1I_5)$ respectively in each run
We run experiments for 15 times and average the results.

The average prediction errors are shown in Table \ref{tab:anti-causal}, where the first three environments are used for training and the last seven are not captured in training with weaker correlation between $V$ and $Y$.
{\bf ERM} and {\bf IRM} achieve the best training performance with respect to their prediction errors on training environments $e_1,e_2,e_3$, while their performances in testing are poor. 
{\bf WDRL} performs worst due to its over pessimism problem. {\bf SAL} achieves nearly uniformly good performance in training environments as well as the testing ones, which validates the effectiveness of our method and proves the excellent generalization ability of SAL.

\begin{table*}[htbp]
	\centering
	\caption{Results of the anti-causal effect experiment. The average prediction errors of 15 runs are reported.}
	\label{tab:anti-causal}
	\vskip 0.05in
	
	\begin{tabular}{|l|c|c|c|c|c|c|c|c|c|c|}
		\hline
		\multicolumn{11}{|c|}{\textbf{Scenario 1: $n_s=5,\ n_v=5$}}\\
		\hline
		$e$&\multicolumn{3}{|c|}{Training environments}&\multicolumn{7}{|c|}{Testing environments}\\
		\hline
		Methods &  $e_1$ & $e_2$ &$e_3$ & $e_4$ &  $e_5$ & $e_6$ &$e_7$ & $e_8$  & $e_9$ & $e_{10}$  \\
		\hline 
		ERM & 0.281 & 0.305 & 0.341 & 0.461 & 0.555 & 0.636& 0.703& 0.733 & 0.765 & 0.824  \\
		LASSO& 0.277& 0.305 & 0.341 & 0.470 & 0.569 & 0.648& 0.722& 0.752 & 0.795 & 0.843\\
		Ridge& \bf 0.258& 0.306 & 0.347 & 0.483 & 0.588 & 0.673& 0.751& 0.783 & 0.828 & 0.879\\
		IRM & 0.287 &\bf 0.293 &\bf 0.329 &\bf 0.345 & 0.382 & 0.420 & 0.444 & 0.461 & 0.478 & 0.504 \\
		WDRL&0.282 & 0.331 & 0.399 & 0.599 & 0.750 & 0.875 & 0.983 & 1.030 & 1.072 & 1.165 \\
		\hline
		SAL &0.324 & 0.329 & 0.331 &  0.358 &\bf 0.381 & \bf 0.403 &\bf 0.425 &\bf 0.435 &\bf 0.446 &\bf 0.458 \\
		\hline
		\multicolumn{11}{|c|}{\textbf{Scenario 2: $n_s=9,\ n_v=1$}}\\
		\hline
		$e$&\multicolumn{3}{|c|}{Training environments}&\multicolumn{7}{|c|}{Testing environments}\\
		\hline
		Methods &  $e_1$ & $e_2$ &$e_3$ & $e_4$ &  $e_5$ & $e_6$ &$e_7$ & $e_8$  & $e_9$ & $e_{10}$  \\
		\hline 
		ERM &\bf 0.272 &\bf  0.280 & 0.298 & 0.526 & 0.362 & 0.411 &
  0.460 &  0.504 &  0.534 &  0.580  \\
  		LASSO& 0.309 & 0.312 & 0.327 & 0.360& 0.397 & 0.425& 0.457  & 0.461 & 0.473 & 0.494\\
		Ridge&0.309 & 0.313 & 0.330 & 0.367 & 0.408 & 0.439& 0.474  & 0.479 & 0.493 & 0.517\\
		IRM & 0.306 &  0.312 & 0.325  & 0.328 & 0.343 &  0.358 & 0.365 & 0.374&  0.377 & 0.394 \\
		WDRL&0.299 & 0.314 & 0.332 & 0.545 & 0.396 & 0.441 &
  0.483 &  0.529 &0.555 &0.596 \\
		\hline
		SAL &0.290 & 0.284 &\bf 0.288 &\bf 0.293 &\bf 0.287 &\bf 0.288 &\bf
  0.287 &\bf  0.290 &\bf 0.284 &\bf  0.294 \\
		\hline
	\end{tabular}
\end{table*}

\subsection{Real Data}
In this section, we test our method on two real-world datasets, and we combine LASSO and Ridge into ERM by setting the coefficient of the regularizer to be $\lambda \geq 0$ due to their similar performances.
\begin{figure*}[!ht]
\small
\vskip -0.1in
     \begin{minipage}{\textwidth}
       \subfigure[$Mean\_Error$ and $Std\_Error$.]{\label{img:house-summary}\includegraphics[width=0.33\textwidth]{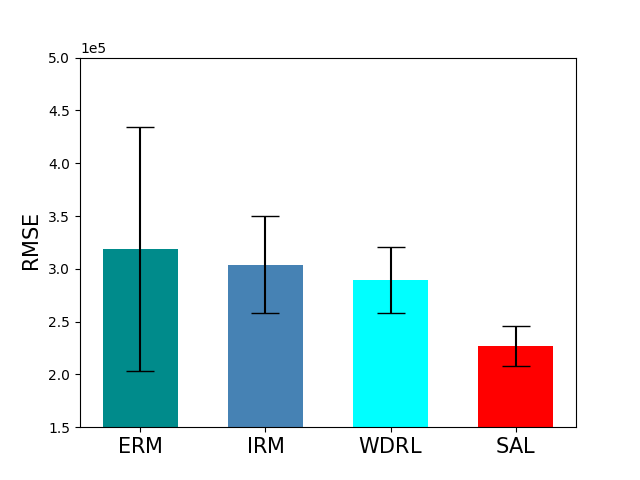}}
		\subfigure[Prediction error with respect to build year.]{\label{img:house-month}\includegraphics[width=0.33\textwidth]{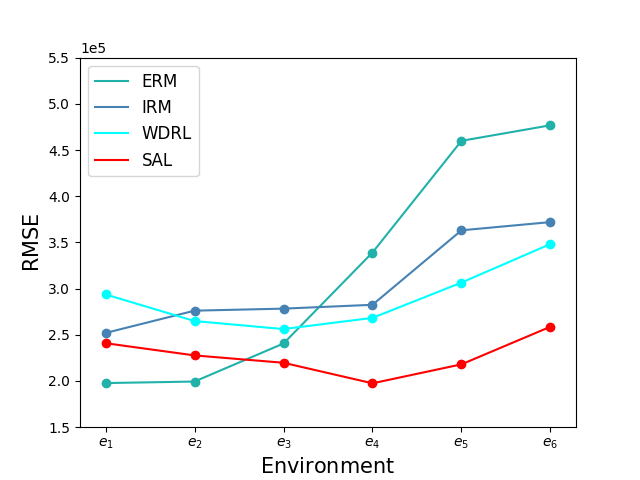}}       
     	\subfigure[Results of the Adult dataset.]{\label{img:adults}
     		\includegraphics[width=0.33\textwidth]{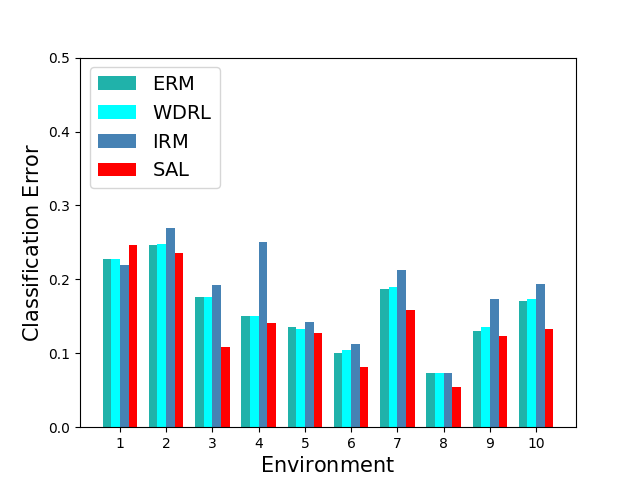}}
     	
     	\caption{Results of the real-world dataset. Figure (a) and (b) are the real regression data and Figure (c) is the Adult dataset.}
     \end{minipage}
     \vskip -0.1in
\end{figure*}

\textbf{Regression}\ \ 
In this experiment, we use a real-world regression dataset (Kaggle) of house sales prices from King County, USA, which includes the houses sold between May 2014 and May 2015 \footnote{https://www.kaggle.com/c/house-prices-advanced-regression-techniques/data}. 
The target variable is the transaction price of the house and each sample contains 17 predictive variables such as the built year of the house, number of bedrooms, and square footage of home, etc. 
We normalize all the predictive covariates to get rid of the influence by their original scales.

To test the stability of different algorithms, we simulate different environments according to the built year of the house.
It is fairly reasonable to assume the correlations between parts of the covariates and the target may vary along time, due to the changing popular style of architecture. 
Specifically, the houses in this dataset were built between $1900\sim 2015$ and we split the dataset into 6 periods, where each period approximately covers a time span of two decades. 
In training, we train all methods on the first and second decade where $built\ year \in [1900,1910)\ and\ [1910,1920)$ respectively and validate on 100 data points sampled from the second period. 

From the results shown in figure \ref{img:house-summary}, we can find that {\bf SAL} achieves not only the smallest $Mean\_Error$ but also the lowest $Std\_Error$ compared with baselines. 
From figure \ref{img:house-month}, we can find that from period 4 and so on, where large distribution shifts occurs, \textbf{ERM} performs poorly and has larger prediction errors. 
\textbf{IRM} performs stably across the first 4 environments but it also fails on the last two, whose distributional shifts are stronger. 
\textbf{WDRL} maintains stable across environments while the mean error is high, which is consistent with our analysis in \ref{exp:toy} that WDRL equally perturbs all covariates and sacrifices accuracy for robustness. 
From figure \ref{img:house-month}, we can find that from period 3 and so on, \textbf{SAL} performs better than ERM, IRM and WDRL, especially when distributional shifts are large. 
In periods 1-2 with slight distributional shift, the SAL method incurs a performance drop compared with IRM and WDRL, while SAL performs much better when larger distributional shifts occur, which is consistent with our intuition that our method sacrifice a little performance in nearly I.I.D. setting for its superior robustness under unknown distribution shifts.

\textbf{Classification}\ \ 
Finally, we validate the effectiveness of our SAL on an income prediction task.
In this task we use the Adult dataset\cite{Dua:2019} which involves predicting personal income levels as above or below \$50,000 per year based on personal details. 
We split the dataset into 10 environments according to demographic attributes, among which distributional shifts might exist.
In training phase, we train all methods on 693 data points from environment 1 and 200 points from the second respectively and validate on 100 points sampled from both.
We normalize all the predictive covariates to get rid of the influence by their original scales. 
In testing phase, we test all methods on the 10 environments and report the mis-classification rate on all environments in figure \ref{img:adults}.
From the results shown in figure \ref{img:adults}, we can find that the \textbf{SAL} outperforms baselines on almost all environments except a slight drop on the first.
However, our SAL outperforms the others in the rest 8 environments where agnostic distributional shifts occur.

%% file: data/7appendix.tex
\begin{proof}[Proof of Theorem 2]
First, we prove that $\mathcal{P} \subseteq \mathcal{P}_0$. $\forall P \in \mathcal{P}$, there exists measure $M_0$ on $\mathcal{Z}\times\mathcal{Z}$ satisfying:
\begin{equation}
    \mathbb{E}_{(z,z')\sim M_0}[c_w(z,z')] \leq \rho
\end{equation}
Note that $c_w$ is optimal if and only if $\min(w^{(i)})=1$ and $\max(w^{(i)})>1$. 
Therefore, we have $\forall z,z' \in \mathcal{Z},\ c(z,z') < c_w(z,z')$. Therefore, we have:
\begin{align}
    W_c(P,P_0) &= \mathop{inf}\limits_{M \in \Pi(P,Q)} \mathbb{E}_{(z,z') \sim M}[c(z,z')]\\
    &\leq \mathbb{E}_{(z,z')\sim M_0}[c(z,z')]\\ 
    &< \mathbb{E}_{(z,z')\sim M_0}[c_w(z,z')] \leq \rho
\end{align}
and therefore $P \in \mathcal{P}_0$ and $\mathcal{P} \subseteq \mathcal{P}_0$. 

Second, we prove that $\exists Q_0 \in \mathcal{P}_0,\ s.t.\ Q_0 \notin \mathcal{P}$ under Assumption 2 and 3.
We have:
\begin{align}
    \mathbb{E}_{(z,z') \sim M_0}[c_w(z,z')] > \mathbb{E}_{(z,z') \sim M_0}[c(z,z')] \geq \rho
\end{align}
and 
\begin{equation}
    \mathbb{E}_{M \in \Pi(P_0,Q_0)-M_0}[c_w(z,z')] > \rho
\end{equation}

which leverages the property that $\|.\|_1$ and $\|.\|_2^2$ are strictly increasing against the absolute value of each covariate of the independent variable. 

For distribution $Q_0$ satisfying Assumption 3, we have:
\begin{align}
    \mathbb{E}_{(z,z') \sim M_0}[c_w(z,z')] &> \rho\\
    \mathbb{E}_{M \in \Pi(P_0,Q_0)-M_0}[c_w(z,z')] &> \rho
\end{align}
and therefore:
\begin{equation}
    \mathop{inf}\limits_{M \in \Pi(P,Q)} \mathbb{E}_{(z,z') \sim M}[c_w(z,z')] > \rho
\end{equation}
which proves that $Q_0 \notin \mathcal{P}$. Therefore, we have $\mathcal{P} \subset \mathcal{P}_0$.

Furthermore, we prove that for the set $U = \{i|w^{(i)}=1\},\ \exists Q_0 \in \mathcal{P}$ that satisfies $W_{c_w}(P_{0\#U},Q_{0\#U})=\rho$ with the help of Assumption 3. 
Assume that distribution $H$ satisfies Assumption 3, we firstly construct a distribution $Q_0$ as following:
\begin{align}
	Q_{0\#U} &= H\\
	\label{equ:proof}
	\forall v \in \mathcal{Z}_{\#U},\ \forall s \in \mathcal{Z}_{\#S}&,\ \ Q_0(s|v) = P_{0\#S}(s)
\end{align}
where $S = \{i|w^{(i)}>1\}$. Since $W_{c_w}(Q_{0\#U}, P_{0\#U}) = \rho$, we have:
\begin{equation}
		\inf_{M \in \Pi(P_{0\#U},Q_{0\#U})} \mathbb{E}_{(z,z') \sim M}[c(z,z')] \leq \rho
\end{equation}
where we refer to the couple minimizing $\mathbb{E}_{(z,z')\sim M_0}[c_w(z,z')]$ as $M_0$. Then we construct joint couple $M$ supported on $\mathcal{Z}\times\mathcal{Z}$, where $M(z,z'),\ z\in\mathcal{Z},\ z'\in\mathcal{Z}$ denotes the probability of transferring $z$ to $z'$. 

Assume $Z=[S,V]$, where $S \in \mathcal{Z}_{\#S}$, $V \in \mathcal{Z}_{\#U}$. $\forall v_1,v_2 \in \mathcal{Z}_{\#U}$, according to equation \ref{equ:proof}, distribution $P_0(S|V=v_1)$ is the same as $Q_0(S|V=v_2)$ and the optimal transportation cost between them is zero. 

For some transportation scheme $\hat{M}$ on $\mathcal{Z}\times\mathcal{Z}$, 
\begin{align}
		&\int_{z \in \mathcal{Z}}\int_{z' \in \mathcal{Z}}c_w(z,z')\hat{M}(z,z') dz dz'\\
		\label{equ:mstar}
		=& \int_{v \in \mathcal{Z}_{\#U}}\int_{v' \in \mathcal{Z}_{\#U}} c_w(v,v')\hat{M}^*(v,v')dvdv'
\end{align}
$\hat{M}^*$ in equation \ref{equ:mstar} denotes the distribution on $\mathcal{Z}_{\#U}\times\mathcal{Z}_{\#U}$. 
Therefore, we have
\begin{align}
		&W_{c_w}(P_0,Q_0) \\
		&= \inf_{M \in \Pi(P_0,Q_0)} \int_{z \in \mathcal{Z}}\int_{z' \in \mathcal{Z}}c_w(z,z')M(z,z') dz dz'\\
		&= \inf_{M \in \Pi(P_{0\#U},Q_{0\#U})}\int_{v \in \mathcal{Z}_{\#U}}\int_{v' \in \mathcal{Z}_{\#U}} c_w(v,v')M(v,v')dvdv'\\
		&= W_{c_w}(P_{0\#U}, Q_{0\#U})= \rho
\end{align}
\end{proof}